\documentclass[a4paper,fleqn]{cas-dc}
\pdfoutput=1

\usepackage[numbers]{natbib}

\usepackage{amssymb}
\usepackage{amsmath}
\usepackage{url}
\usepackage{soul}
\usepackage{times}
\usepackage{array}
\usepackage{bbding}
\usepackage{amsfonts}
\usepackage{graphicx}
\usepackage{textcomp}
\usepackage{multirow}
\usepackage{booktabs}
\usepackage{makecell}
\usepackage{algorithm}
\usepackage{algorithmic}
\usepackage[utf8]{inputenc}
\usepackage[small]{caption}
\usepackage[switch]{lineno}

\usepackage{cleveref}

\def\tsc#1{\csdef{#1}{\textsc{\lowercase{#1}}\xspace}}
\tsc{WGM}
\tsc{QE}

\begin{document}
\let\WriteBookmarks\relax
\def\floatpagepagefraction{1}
\def\textpagefraction{.001}
\let\printorcid\relax

\shorttitle{A Survey on RGB, 3D, and Multimodal Approaches for Unsupervised Industrial Image Anomaly Detection} 

\shortauthors{Yuxuan Lin et al.}

\title [mode = title]{A Survey on RGB, 3D, and Multimodal Approaches for Unsupervised Industrial Image Anomaly Detection}  



\author[1]{Yuxuan Lin}
\ead{yuxuanlin24@m.fudan.edu.cn}

\author[2]{Yang Chang}
\ead{ychang24@m.fudan.edu.cn}

\author[2]{Xuan Tong}
\ead{xtong23@m.fudan.edu.cn}

\author[2]{Jiawen Yu}
\ead{jwyu23@m.fudan.edu.cn}

\author[3]{Antonio Liotta}
\ead{antonio.liotta@unibz.it}

\author[2]{Guofan Huang}
\ead{gfhuang24@m.fudan.edu.cn}

\author[4]{Wei Song}
\ead{wsong@shou.edu.cn}

\author[5]{Deyu Zeng}
\ead{deyuzeng@szu.edu.cn}

\author[5]{Zongze Wu}
\ead{zzwu@szu.edu.cn}

\author[2]{Yan Wang}
\cormark[1]
\ead{yanwang19@fudan.edu.cn}

\author[1,2]{Wenqiang Zhang}
\cormark[1]
\ead{wqzhang@fudan.edu.cn}

\address[1]{Shanghai Key Lab of Intelligent Information Processing, School of Computer Science, Fudan University, Shanghai 200438, China}
\address[2]{Shanghai Engineering Research Center of AI \& Robotics, Academy for Engineering \& Technology, Fudan University, Shanghai 200438, China.}
\address[3]{Faculty of Computer Science, Free University of Bozen-Bolzano, Bolzano 39100, Italy}
\address[4]{College of Information Technology, Shanghai Ocean University, Shanghai 201306, China}
\address[5]{College of Mechatronics and Control Engineering, Shenzhen University, Shenzhen 518060, China}

\cortext[cor1]{Corresponding author}

\begin{abstract}
In the advancement of industrial informatization, unsupervised anomaly detection technology effectively overcomes the scarcity of abnormal samples and significantly enhances the automation and reliability of smart manufacturing. As an important branch, industrial image anomaly detection focuses on automatically identifying visual anomalies in industrial scenarios (such as product surface defects, assembly errors, and equipment appearance anomalies) through computer vision techniques. With the rapid development of Unsupervised industrial Image Anomaly Detection (UIAD), excellent detection performance has been achieved not only in RGB setting but also in 3D and multimodal (RGB and 3D) settings. However, existing surveys primarily focus on UIAD tasks in RGB setting, with little discussion in 3D and multimodal settings. To address this gap, this artical provides a comprehensive review of UIAD tasks in the three modal settings. Specifically, we first introduce the task concept and process of UIAD. We then overview the research on UIAD in three modal settings (RGB, 3D, and multimodal), including datasets and methods, and review multimodal feature fusion strategies in multimodal setting. Finally, we summarize the main challenges faced by UIAD tasks in the three modal settings, and offer insights into future development directions, aiming to provide researchers with a comprehensive reference and offer new perspectives for the advancement of industrial informatization. Corresponding resources are available at \url{https://github.com/Sunny5250/Awesome-Multi-Setting-UIAD}.
\end{abstract}





\maketitle


\section{Introduction}
Anomaly detection plays a key role in the stable operation, fault prevention, loss reduction and efficiency improvement of industrial systems. Traditional manual detection methods rely on operator observation and analysis of captured images. This approach is usually time-intensive and its efficiency is limited by human processing power and attention. When facing large volumes of data or complex industrial environments, the efficiency and accuracy of manual detection often significantly decrease. With the development of deep learning algorithms, detection methods based on deep learning are gradually replacing manual detection, and they have the advantages of being more efficient, more accurate, and more robust. With the continuous advancement of industrial informatization, the amount of information in the production process has increased dramatically, and traditional detection methods can no longer meet the requirements for real-time and accurate detection.

\begin{figure}[t]
    \centering
    \includegraphics[width=1\columnwidth]{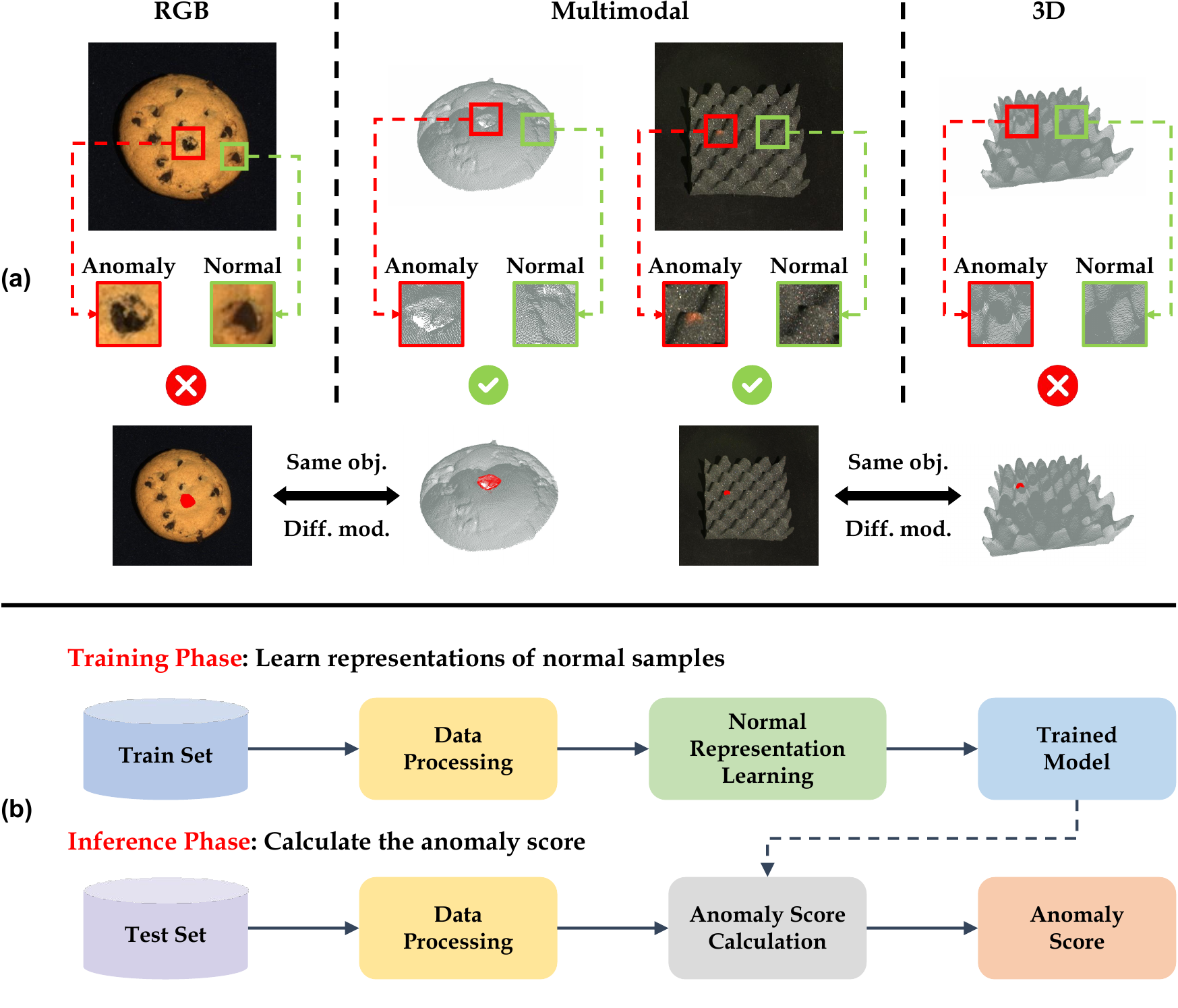}
    \caption{Examples of some classes in the MVTec 3D-AD dataset and the unsupervised industrial anomaly detection (UIAD) pipeline. (a) We take RGB, 3D and multimodal samples as examples, and find that different modals of the same object can compensate for the information limitations of a single modal, enabling the detection of more types of anomalies. (b) The training and inference phases of UIAD.}
    \label{motivation}
\end{figure}

Among existing detection methods based on deep learning, unsupervised methods are more widespread than supervised methods. There are three reasons: (1) In actual industrial environments, abnormal situations are relatively rare, so it is difficult to obtain a large number of labeled abnormal samples for supervised learning. (2) The types and forms of industrial anomalies may be very diverse, and it is difficult to predefine or label all possible anomaly types. (3) Unsupervised learning methods identify anomalies by learning the distribution of normal data, which enables them to identify new types of anomalies that do not appear in the training data and have better generalization capabilities. Unsupervised learning methods can learn the data distribution under normal conditions without or only a small amount of labeled data, and then identify abnormal patterns that deviate from this distribution. They are suitable for industrial detection scenarios where abnormal samples are scarce or have high diversity.

Fig. \ref{motivation} shows why we need different modal information for anomaly detection and the UIAD process. We can find that the same type of anomaly exhibits significant differences under different modal information. Combining different modal information helps us detect multiple types of anomalies more comprehensively. For UIAD, during the training phase, the model receives processed single-modal or multimodal data and learns representations of normal samples. In the inference phase, the learned model receives the processed single-modal or multimodal data to be detected and calculates the final anomaly score.

Industrial anomaly detection involves various modal inputs, such as images, videos, time series, etc., with different modal inputs corresponding to different tasks. We focus on image anomaly detection, which is one of the widely discussed tasks in computer vision. Unlike tasks such as video and time series anomaly detection, image anomaly detection detects anomalies on the surface or structure of objects by extracting visual features from images. With the increasing maturity of RGB and 3D imaging technologies, industrial vision detection systems are gradually transitioning from single-modal to multimodal. RGB images, due to their rich color and texture information, remain the primary data source in most industrial applications. However, 3D data (such as point clouds and depth maps) can provide additional spatial information, enhancing the accuracy and robustness of detection. Moreover, multimodal approaches combine different types of visual modal data (such as combining RGB and 3D point clouds), enabling more comprehensive detection of various types of industrial anomalies and addressing potential information gaps in single-modal approaches.

Single-modal image anomaly detection usually uses RGB images as the detection object, so the information contained in RGB graphics becomes the only criterion for anomalies. However, single-modal information may not be able to fully reflect the operating conditions of complex industrial systems, especially when facing highly complex and changing industrial environments. Currently, anomaly detection methods based on RGB images are mainstream in the industrial field. Inspired by medical anomaly detection, the industrial field extends the anomaly detection task to other modal information. For example, 3D point clouds obtained through sensors are used for some specific tasks in UIAD.

\begin{figure}[t]
    \centering
    \includegraphics[width=1\columnwidth]{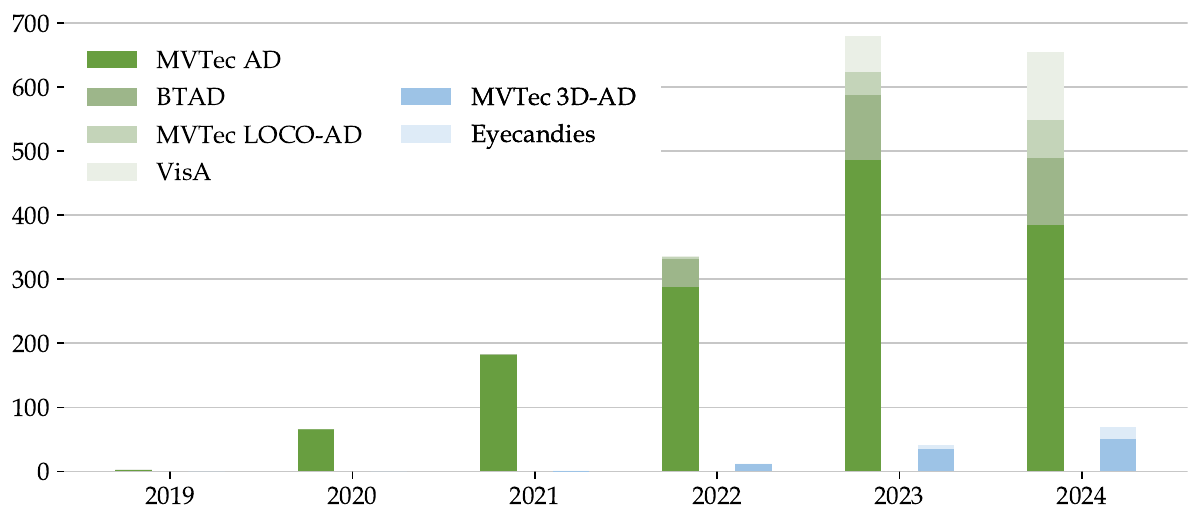}
    \caption{The number of citations for some respective RGB and multimodal datasets over time (green for RGB, blue for multimodal). Among RGB datasets, MVTec AD has the highest citation ratio, which has a profound impact on the field of RGB anomaly detection. It is also evident that the number of citations for multimodal datasets has increased year by year since they were proposed, indicating that multimodal information is receiving more and more attention. Citation counts were obtained in Google Scholar.}
    \label{cite-num}
\end{figure}

Multimodal anomaly detection, which has recently gained increasing attention, uses multiple input information as a criterion and can provide better detection accuracy for anomalous areas. We count the number of citations for some RGB and multimodal datasets over time, as shown in Fig. \ref{cite-num}. Multimodal anomaly detection methods can capture system conditions more comprehensively by integrating multiple modal information (such as RGB images, 3D point clouds, infrared images, etc.). This integration not only improves the accuracy of the anomaly detection algorithm but also enhances the sensitivity of the algorithm to subtle changes. Taking the cookie class in the MVTec 3D-AD dataset \cite{bergmann2021mvtec3d} as an example, the chocolate and hole anomalies on the cookie surface are very similar in the RGB image, but are easier to distinguish in the 3D point cloud. In addition, using multimodal information as a criterion helps improve the robustness of anomaly detection algorithms. In complex industrial environments, certain modal information may be disturbed or unavailable due to various factors. Multimodal anomaly detection methods can supplement or verify the information with other modal information sources, thus maintaining the stability of the detection performance. According to BTF \cite{horwitz2023back}, we can know that 3D modal is often needed because it can be used to make up for the shortcomings of RGB modal. At the same time, the two modals can complement each other, and fusion of different modals can pay attention to more types of abnormal situations and improve detection accuracy. The release of the MVTec 3D-AD dataset advances the development of multimodal anomaly detection methods, enabling industrial informatization systems to maintain higher stability and efficiency in environments with high complexity and multi-source data.

\begin{figure*}[ht]
    \centering
    \includegraphics[width=1\textwidth]{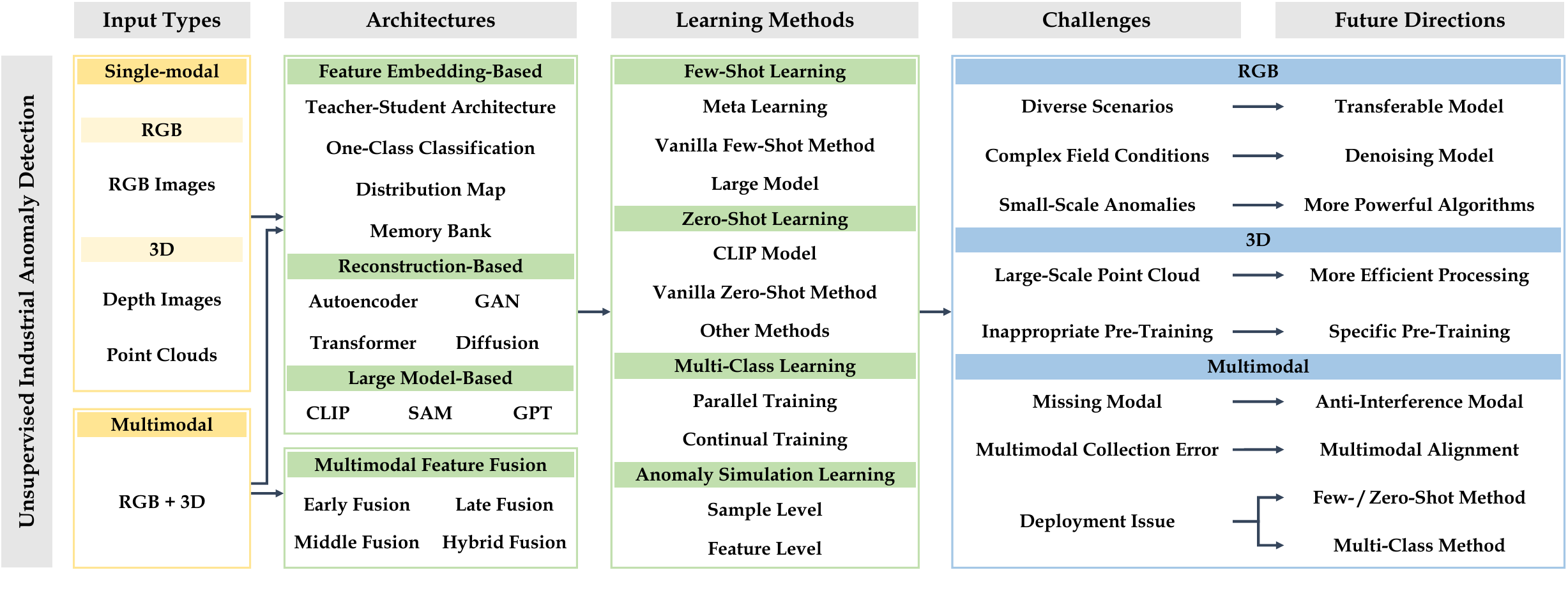}
    \caption{\textbf{Roadmap of RGB, 3D, multimodal Unsupervised Industrial Anomaly Detection (UIAD).} We conduct an in-depth analysis of RGB, 3D, and multimodal UIAD, summarizing existing UIAD methods from the perspectives of input types, architectures and learning methods. We identify the commonalities and distinctions of UIAD methods in different settings. Based on the above analysis, we discuss the challenges faced by UIAD methods in different settings and corresponding potential future research directions.}
    \label{road-map}
\end{figure*}

\begin{table*}
    \centering
    \renewcommand\arraystretch{1.0}
    \resizebox{0.9\linewidth}{!}{
    \begin{tabular}{lcccccccccccccccccccc}
        \toprule
        \multirow{2}{*}{\textbf{Survey}} & \multirow{2}{*}{\textbf{Year}} & \multicolumn{9}{c}{\textbf{RGB UIAD}} & \multicolumn{5}{c}{\textbf{3D UIAD}} & \multicolumn{5}{c}{\textbf{Multimodal UIAD}} \\
        \cmidrule(lr{0pt}){3-11} \cmidrule(lr{0pt}){12-16} \cmidrule(lr{0pt}){17-21}
        & & DS & MT & LM & FS & ZS & MC & AS & CL & FT & DS & MT & AS & CL & FT & DS & MT & MF & CL & FT \\
        \midrule
        Tao \cite{tao2022deep} & 2022 & \Checkmark & \Checkmark & \XSolidBrush & \XSolidBrush & \XSolidBrush & \XSolidBrush & \Checkmark & \Checkmark & \Checkmark & \XSolidBrush & \XSolidBrush & \XSolidBrush & \XSolidBrush & \XSolidBrush & \Cross & \XSolidBrush & \XSolidBrush & \XSolidBrush & \Cross \\
        Cui \cite{cui2023survey} & 2023 & \Checkmark & \Checkmark & \Cross & \XSolidBrush & \XSolidBrush & \XSolidBrush & \Checkmark & \Checkmark & \Checkmark & \XSolidBrush & \XSolidBrush & \XSolidBrush & \XSolidBrush & \XSolidBrush & \XSolidBrush & \XSolidBrush & \XSolidBrush & \XSolidBrush & \XSolidBrush \\
        Diers \cite{diers2023survey} & 2023 & \Checkmark & \Checkmark & \Cross & \Checkmark & \Checkmark & \XSolidBrush & \Checkmark & \Checkmark & \Checkmark & \XSolidBrush & \XSolidBrush & \XSolidBrush & \XSolidBrush & \XSolidBrush & \Cross & \XSolidBrush & \XSolidBrush & \XSolidBrush & \XSolidBrush \\
        Liu \cite{liu2024deep} & 2024 & \Checkmark & \Checkmark & \Cross & \Checkmark & \Checkmark & \XSolidBrush & \Checkmark & \Checkmark & \Checkmark & \XSolidBrush & \Cross & \XSolidBrush & \XSolidBrush & \XSolidBrush & \Cross & \Cross & \XSolidBrush & \Cross & \Cross \\
        Cao \cite{cao2024survey} & 2024 & \Checkmark & \Checkmark & \Cross & \Checkmark & \Checkmark & \XSolidBrush & \Cross & \Checkmark & \Checkmark & \Cross & \Cross & \XSolidBrush & \Cross & \XSolidBrush & \Cross & \Checkmark & \XSolidBrush & \XSolidBrush & \XSolidBrush \\
        \midrule
        Ours & 2024 & \Checkmark & \Checkmark & \Checkmark & \Checkmark & \Checkmark & \Checkmark & \Checkmark & \Checkmark & \Checkmark & \Checkmark & \Checkmark & \Checkmark & \Checkmark & \Checkmark & \Checkmark & \Checkmark & \Checkmark & \Checkmark & \Checkmark \\
        \bottomrule
    \end{tabular}}
    \caption{Comparison of related surveys and ours for UIAD (Only include works in unsupervised industrial settings). \Checkmark, \Cross and \XSolidBrush represent Comprehensive Summary, Briefly Mentioned and Not Mentioned, respectively. DS, MT, LM, FS, ZS, MC, AS, MF, CL and FT represent Dataset, Method, Large Model-based method, Few-Shot method, Zero-Shot method, Multi-Class method, Anomaly Simulation method, Multimodal Fusion method, Challenge and Future, respectively.}
    \label{tab:survey}
\end{table*}

Although the field of multimodal anomaly detection has made significant progress in recent years, especially driven by deep learning and sensor technologies, a comprehensive overview summarizing the progress and its practical applications in industrial systems is currently not available. Table \ref{tab:survey} shows the advantages of our survey compared with previous surveys of industrial anomaly detection. We provide a more comprehensive summary of methods and datasets in the fields of pure RGB, pure 3D, and multimodal anomaly detection, and discuss relevant research directions based on the needs of the industrial manufacturing. From the perspective of modal settings, we introduce detection tasks in different modal settings to meet the varying needs of different production lines for different types of data. For example, production lines for low-cost, simple products may only require RGB images to complete the detection process. In contrast, production lines for high-cost, complex products may need to combine high-resolution RGB images with 3D point cloud data to complete the detection process. To the best of our knowledge, this survey is the first to systematically classify and deeply review UIAD in different settings, with special attention to the starting point of their design and the challenges they face when applied in practice. The framework roadmap of this survey is shown in Fig. \ref{road-map}. The paper mainly makes the following contributions:

\begin{itemize}
\item To the best of our knowledge, we are the first to present a comprehensive survey for the unsupervised industrial image anomaly detection task in different settings (RGB, 3D and multimodal).
\item We systematically categorize existing UIAD methods according to different paradigms and summarize current optimization directions in industrial manufacturing (few-/zero-shot methods, multi-class methods). Additionally, we summarize common datasets and evaluation metrics, which can aid researchers in better understanding task orientations and focusing more intently on their areas of interest.
\item We classify and summarize the modal information fusion strategies in recent multimodal UIAD methods (early, middle, late, hybrid fusion) to facilitate researchers in innovating based on existing fusion methods.
\item We highlight the major issues and challenges that existing research faces in different settings (RGB, 3D and multimodal) in terms of data processing, methodologies, and applications. Guided by proposed challenges, we propose potential research directions for future work, aiming to provide new insights for unsupervised industrial anomaly detection tasks.
\end{itemize}

\begin{table*}
    \centering
    \renewcommand\arraystretch{1.0}
    \resizebox{\linewidth}{!}{
    \begin{tabular}{clcccccccccc}
        \toprule
        & \multirow{2}{*}{\textbf{Datasets}} & \multirow{2}{*}{\textbf{Year}} & \multirow{2}{*}{\textbf{Type}} & \multicolumn{5}{c}{\textbf{Number}} & \multirow{2}{*}{\textbf{Class}} & \multirow{2}{*}{\textbf{Anomaly Type}} & \multirow{2}{*}{\textbf{Modal Type}} \\
        \cmidrule(lr{0pt}){5-9}
        & & & & \# Train & \# Test (good) & \# Test (anomaly) & \# Val & Total \\
        \midrule
        \multirow{13}*{\rotatebox{90}{RGB}}
        & MVTec AD \cite{bergmann2019mvtec,bergmann2021mvtec} & 2019 & Real & 3,629 & 467 & 1,258 & - & 5,354 & 15 & 73 & RGB \\
        & BTAD \cite{mishra2021vt} & 2021 & Real & 1,799 & 451 & 290 & - & 2,540 & 3 & - & RGB \\
        & MPDD \cite{jezek2021deep} & 2021 & Real & 888 & 176 & 282 & - & 1,346 & 6 & - & RGB \\
        & MVTec LOCO-AD \cite{bergmann2022beyond} & 2022 & Real & 1,772 & 575 & 993 & 304 & 3,644 & 5 & 89 & RGB \\
        & VisA \cite{zou2022spot} & 2022 & Real & 9,621 & 0 & 1,200 & - & 10,821 & 12 & - & RGB \\
        & GoodsAD \cite{zhang2024pku} & 2023 & Real & 3,136 & 1,328 & 1,660 & - & 6,124 & 6 & - & RGB \\
        & MSC-AD \cite{zhao2023msc} & 2023 & Real & 6,480 & 2,160 & 1,080 & - & 9,720 & 12 & 5 & RGB \\
        & CID \cite{zhang2024catenary} & 2024 & Real & 3,900 & 33 & 360 & - & 	4,293 & 1 & 6 & RGB \\
        & Real-IAD \cite{wang2024real} & 2024 & Real & 72,840 & 0 & 78,210 & - & 151,050 & 30 & 8 & RGB \\
        & RAD \cite{cheng2024rad} & 2024 & Real & 213 & 73 & 1,224 & - & 1,510 & 4 & - & RGB \\
        & MIAD \cite{bao2023miad} & 2023 & Synthetic & 70,000 & 17,500 & 17,500 & - & 105,000 & 7 & 13 & RGB \\
        & MAD-Sim \cite{zhou2024pad} & 2023 & Synthetic & 4,200 & 638 & 4,951 & - & 9,789 & 20 & 3 & RGB \\
        & DTD-Synthetic \cite{aota2023zero} & 2024 & Synthetic & 1,200 & 357 & 947 & - & 2,504 & 12 & - & RGB \\
        \midrule
        \multirow{2}*{\rotatebox{90}{3D}}
        & Real3D-AD \cite{liu2023real3d} & 2024 & Real & 48 & 604 & 602 & - & 1,254 & 12 & 3 & Point cloud \\
        & Anomaly-ShapeNet \cite{li2023towards} & 2024 & Synthetic & 208 & 780 & 943 & - & 1,931 & 50 & 7 & Point cloud \\
        \midrule
        \multirow{3}*{\rotatebox{90}{\makecell{Multi-\\modal}}}
        & MVTec 3D-AD \cite{bergmann2021mvtec3d} & 2021 & Real & 2,656 & 294 & 948 & 294 & 4,147 & 10 & 41 & RGB \& Point cloud \\
        & PD-REAL \cite{qin2023image} & 2023 & Real & 2,399 & 300 & 530 & 300 & 3,529 & 15 & 6 & RGB \& Point cloud \\
        & Eyecandies \cite{bonfiglioli2022eyecandies} & 2022 & Synthetic & 10,000 & 2,250 & 2,250 & 1,000 & 15,500 & 10 & - & RGB \& Depth \\
        \bottomrule
    \end{tabular}}
    \caption{Comparison of RGB, 3D, multimodal datasets for UIAD.}
    \label{tab:datasets}
\end{table*}

\section{Preliminaries}

\subsection{Unsupervised Anomaly Detection}

Unsupervised anomaly detection refers to using unsupervised methods to detect and localize anomaly patterns or outliers in samples through one or more modals of data. The unsupervised industrial image anomaly detection discussed in this artical uses normal sample images as training data, allowing the model to learn the features of the normal pattern in the images. First, we use a training set $\mathcal{X}=\left\{x_{1}, x_{2}, \ldots, x_{N}\right\}$ containing only normal data samples to train an anomaly detection model $f$ that learns representations of normal patterns. The training goal is to find the optimal model parameters so that the model $f$ can effectively represent the distribution of normal data, which can be expressed by the following optimization problem:

\begin{equation}
  \theta^{*}=\arg \min _{\theta} \mathcal{L} \left(f \left( x \in \mathcal{X};\theta \right)\right)
\end{equation}

\noindent where $\theta$ represents the model parameters, $\theta^{*}$ represents the optimal model parameters, and $\mathcal{L}$ is the loss function for the model $f$ to learn normal data representations.
In the testing phase, for the sample $x_{test}$ to be detected, we use the trained model $f$ to perform anomaly detection and localization:

\begin{equation}
  s_{test}=f \left( x_{test};\theta^{*} \right)
\end{equation}

\noindent where $s_{test}$ is usually the anomaly score of the test sample $x_{\text{test}}$, reflecting the degree of deviationof the sample from the learned normal data distribution. The higher the anomaly score, the more likely the sample contains anomalies. Based on the ground truth of the test samples, we can compute evaluation metrics to evaluate the model:

\begin{equation}
  \mathcal{M} = \text{Evaluate} \left( f \left( x_{test};\theta^{*} \right), y_{test} \right)
\end{equation}

If the input data modals are more than one, we also need to use multimodal fusion methods. Multimodal feature fusion aims to merge information from different modals into a unified representation. This can be achieved through various methods, such as simple concatenation, feature transformation, or more complex neural network architectures.

\subsection{Datasets}

We summarize the main UIAD datasets in this survey in Table \ref{tab:datasets}.

\subsubsection{RGB UIAD datasets}

For RGB UIAD datasets, the earliest one is MVTec AD \cite{bergmann2019mvtec}. MVTec AD simulates real industrial production scenarios, with only normal samples in the training set and both normal and abnormal samples in the test set, primarily used for unsupervised anomaly detection. MVTec LOCO-AD \cite{bergmann2022beyond} introduces logical anomaly types, such as misplacement or missing objects. VisA \cite{zou2022spot} introduces complex structure objects and changes in pose/position. MSC-AD \cite{zhao2023msc} focuses on detecting tiny anomalies across multiple scenes, collecting real-world large-scale industrial casting surface anomalies under varying brightness and resolution. Real-IAD \cite{wang2024real} collects large-scale, multi-view samples to better distinguish the performance of methods.

\subsubsection{3D UIAD datasets}

For 3D UIAD datasets, point cloud data is currently used. Real3D-AD \cite{liu2023real3d} is the first high-resolution real point cloud anomaly detection dataset. Anomaly-ShapeNet \cite{li2023towards} is the first high-quality synthetic point cloud anomaly dataset.

\subsubsection{Multimodal UIAD datasets}

For multimodal UIAD datasets, the most commonly used is MVTec 3D-AD \cite{bergmann2021mvtec3d}, the first real multimodal dataset for multimodal UIAD, containing RGB images and corresponding 3D point cloud data. Eyecandies \cite{bonfiglioli2022eyecandies} is the first synthetic multimodal dataset for multimodal UIAD. PD-REAL \cite{qin2023image} adopts a low-cost dataset construction approach, which has the advantages of scalability and easy variable control.

\subsection{Evaluation Metrics}

In the task of image anomaly detection, the selection of evaluation metrics is crucial. We divide the commonly used evaluation metrics into three categories, namely image-level metrics, pixel-level metrics and instance-level metrics. Before introducing the various metrics, we first present the fundamental elements for metric calculation: $TP$ (True Positive) refers to the number of images/pixels that are anomalous and correctly predicted as anomalous, $FN$ (False Negative) refers to the number of images/pixels that are anomalous but incorrectly predicted as normal, $FP$ (False Positive) refers to the number of images/pixels that are normal but incorrectly predicted as anomalous, and $TN$ (True Negative) refers to the number of images/pixels that are normal and correctly predicted as normal. We summarize the main metrics in Table \ref{tab:metrics}.

\begin{table*}
    \centering
    \renewcommand\arraystretch{1.0}
    \resizebox{\linewidth}{!}{
    \begin{tabular}{cllll}
        \toprule
        \textbf{Level} & \textbf{Abbr.} & \textbf{Metric} & \textbf{Formula} & \textbf{Remarks} \\
        \midrule
        \multirow{4}*{-}
        & $P$ $\uparrow$ & Precision & $P = TP / \left(TP + FP\right)$ & - \\
        & $R$ $\uparrow$ & Recall & $R = TP / \left(TP + FN\right)$ & - \\
        & $TPR$ $\uparrow$ & True positive rate & $TPR = TP / \left(TP + FN\right)$ & - \\
        & $FPR$ $\downarrow$ & False positive rate & $FPR = FP / \left(FP + TN\right)$ & - \\
        \midrule
        \multirow{2}*{Image}
        & $I\text{-}AUROC$ $\uparrow$ & Image-level area under the ROC curve & $I\text{-}AUROC = \int_{0}^{1} \left( TPR \right) d \left( FPR \right)$ & - \\
        & $F1$ score $\uparrow$ & F1 score / Balanced F-score & $F1 = 2 \left( P \times R \right) / \left( P + R \right)$ & - \\
        \midrule
        \multirow{7}*{Pixel}
        & $P\text{-}AUROC$ $\uparrow$ & Pixel-level area under the ROC curve & $P\text{-}AUROC = \int_{0}^{1} \left( TPR \right) d \left( FPR \right)$ & - \\
        & $AUPRO$ $\uparrow$ & Area under per-region overlap curve & $AUPRO = \text{Norm}\left(\int_{0}^{fl} \left( \left( 1/K \right) \sum_{k=1}^{K} \left( \left| P \cap C_{k} \right| / \left| C_{k} \right| \right) \right) d \left( FPR \right)\right)$ & \vtop{\hbox{\strut FPR limit ($fl$)}\hbox{\strut Total number of ground truth components ($K$)}\hbox{\strut Binary prediction ($P$)}\hbox{\strut Each connected component ($C_{k}$)}} \\
        & $IoU$ $\uparrow$ & Intersection over union & $IoU = TP / \left(FN + FP + TP\right)$ & - \\
        & $P\text{-}AP$ $\uparrow$ & Pixel-level average precision & $P\text{-}AP = \int_{0}^{1} \left( P \right) d \left( R \right)$ & Also known as area under precision-recall (AUPR) \\
        \midrule
        \multirow{1}*{Instance}
        & $IAP$ $\uparrow$ & Instance average precision & $IAP = \left( 1/N \right) \sum_{i=1}^{N} P\left( i \right) $ & Total number of instances ($N$) \\
        \bottomrule
    \end{tabular}}
    \caption{A summary of metrics used for UIAD.}
    \label{tab:metrics}
\end{table*}

\subsubsection{Image-Level Metrics}

This kind of metric is used to determine whether the whole image contains anomalies. It does not address the specific location or size of the anomaly, but simply marks the whole image as normal or abnormal. For example, if there are any anomaly areas in an image, the entire image is considered abnormal. Commonly used image-level metrics include image-level area under the receiver operating characteristic curve ($I\text{-}AUROC$).

\subsubsection{Pixel-Level Metrics}

This kind of metric is used to evaluate the abnormal status of each pixel in the image. It is suitable for tasks that require precise determination of the location of anomaly areas. By evaluating whether each pixel is correctly labeled as normal or abnormal, pixel-level metrics can accurately measure the performance of models at a detailed level. Commonly used pixel-level metrics include pixel-level area under the receiver operating characteristic curve ($P\text{-}AUROC$), area under the per-region overlap curve ($AUPRO$), intersection over union ($IoU$) and pixel-level average precision ($P\text{-}AP$).

\subsubsection{Instance-Level Metrics}

According to DeSTSeg \cite{zhang2023destseg}, for UIAD tasks, each anomaly instance is defined as a maximally connected ground truth area (e.g., all pixels of an anomaly area). Instance-level evaluation focuses on correctly detecting and localizing an instance (such as an anomaly area on a workpiece), either wholly or partially, rather than every individual pixel. An anomaly instance is considered detected if and only if more than 50\% of the area’s pixels are predicted as anomaly. Common instance-level metrics include instance average precision ($IAP$) and $IAP@k$ (the precision at $recall = k\%$).

\begin{table*}\footnotesize
    \centering
    \renewcommand\arraystretch{1.0}
    \resizebox{\linewidth}{!}{
    \begin{tabular}{lclp{11cm}}
        \toprule
        \textbf{Methods} & \textbf{Year} & \textbf{Paradigm} & \textbf{Highlight} \\
        \midrule
        US \cite{bergmann2020uninformed} & 2020 & Teacher-student architecture & Used the teacher-student network architecture for the first time in anomaly detection. \\
        MKD \cite{salehi2021multiresolution} & 2021 & Teacher-student architecture & Proposed multiresolution knowledge distillation. \\
        GP \cite{wang2021glancing} & 2021 & Teacher-student architecture & Combined local and global features and adopted a double-head comparison mechanism. \\
        RD4AD \cite{deng2022anomaly} & 2022 & Teacher-student architecture & Proposed a reverse distillation method and eliminated redundant features across multiple scales. \\
        PFM \cite{wan2022unsupervised} & 2023 & Teacher-student architecture & Proposed pre-trained feature mapping methods to detect and locate anomalies. \\
        MemKD \cite{gu2023remembering} & 2023 & Teacher-student architecture & Used memory modules and a normality embedding learning strategy to alleviate the "normality forgetting" problem of student networks. \\
        DeSTSeg \cite{zhang2023destseg} & 2023 & Teacher-student architecture & Proposed a denoising student encoder-decoder and a segmentation network to fuse multi-scale features for anomaly detection. \\
        EfficientAD \cite{batzner2024efficientad} & 2024 & Teacher-student architecture & Implemented an efficient network for real-time inference with millisecond-level latency. \\
        EMMFRKD \cite{tong2024enhanced} & 2024 & Teacher-student architecture & Proposed enhanced mutual mapping feature fusion, coordinated attention mechanism and single-category embedding memory bank. \\
        AEKD \cite{wu2024aekd} & 2024 & Teacher-student architecture & Proposed asymmetric structure, multi-scale features fusion and different data flows to alleviate the generalization issue of the similarity between student and teacher. \\
        FCACDL \cite{zhang2024feature} & 2024 & Teacher-student architecture & Introduced multi-view consistency constraint, central feature constraint and encoding consistency. \\
        DMDD \cite{liu2024dual} & 2024 & Teacher-student architecture & Proposed the decoupled teacher-student network and dual modeling distillation mechanism. \\
        \midrule
        CutPaste \cite{li2021cutpaste} & 2021 & One-class classification & Proposed a new cut-and-paste anomaly simulation method. \\
        SimpleNet \cite{liu2023simplenet} & 2023 & One-class classification & Simulated anomalies at the feature level rather than at the image level. \\
        ADShift \cite{cao2023anomaly} & 2023 & One-class classification & Minimized the distribution gap between in-/out-of-distribution in both training and inference phases to improved the performance of anomaly detection under distribution shift. \\
        DS2 \cite{he2024learning} & 2024 & One-class classification & Proposed a two-stage, dense pre-training model for image anomaly localization task. \\
        GeneralAD \cite{strater2024generalad} & 2024 & One-class classification & Introduced a self-supervised anomaly feature generation and a cross-patch attention discriminator. \\
        GLASS \cite{chen2024unified} & 2024 & One-class classification & Combined the global and local anomaly synthesis strategy. \\
        \midrule
        FastFlow \cite{yu2021fastflow} & 2021 & Distribution map & Simulated global and local distribution by alternately stacking large and small convolution kernels. \\
        DifferNet \cite{rudolph2021same} & 2021 & Distribution map & Simulated global and local distribution by alternately stacking large and small convolution kernels. \\
        CFLOW-AD \cite{gudovskiy2022cflow} & 2022 & Distribution map & Introduced position encoder into the conditional normalized flow network. \\
        CS-Flow \cite{rudolph2022fully} & 2022 & Distribution map & Used multi-scale images and features in the normalization flow module. \\
        CDO \cite{cao2023collaborative} & 2023 & Distribution map & Used collaborative optimization to reduce the overlap between normal and anomaly distributions to alleviate the overgeneralization problem. \\
        PyramidFlow \cite{lei2023pyramidflow} & 2023 & Distribution map & Proposed invertible pyramids and pyramid coupling blocks for multi-scale fusion and mapping. \\
        SLAD \cite{xu2023fascinating} & 2023 & Distribution map & Introduced supervised signals based on scale learning. \\
        MSFlow \cite{zhou2024msflow} & 2024 & Distribution map & Proposed asymmetrical parallel flows followed by a fusion flow to exchange multi-scale perceptions. \\
        AttentDifferNet \cite{simoes2024attention} & 2024 & Distribution map & Combined normalizing flow with attention mechanisms. \\
        \midrule
        PaDiM \cite{defard2021padim} & 2021 & Memory bank & Used a pre-trained CNN and multivariate Gaussian distribution to model the features of normal patches. \\
        PatchCore \cite{roth2022towards} & 2022 & Memory bank & Introduced a greedy coreset subsampling method to reduce the redundancy of the memory bank. \\
        CFA \cite{lee2022cfa} & 2022 & Memory bank & Clustered normal features by constructing coupled hyperspheres. \\
        DMAD \cite{liu2023diversity} & 2023 & Memory bank & Compressed input feature embeddings into the nearest single memory item in memory banks, replacing the combination of multiple memory items. \\
        PNI \cite{bae2023pni} & 2023 & Memory bank & Introduced positional information and neighborhood information in the memory bank module. \\
        GraphCore \cite{xie2023pushing} & 2023 & Memory bank & Introduced a vision isometric invariant graph neural network to build memory banks in few-shot settings. \\
        InReaCh \cite{mcintosh2023inter} & 2023 & Memory bank & Built memory banks by extracting high-confidence nominal patches from training data in channels with high spans and low spread. \\
        ReconFA \cite{zuo2024reconstruction} & 2024 & Memory bank & Adapted the normal features to target domain to achieve more compact memory banks. \\
        ReConPatch \cite{hyun2024reconpatch} & 2024 & Memory bank & Proposed a similarity-based linear transformation to optimize the build of the memory bank. \\
        \bottomrule
            \end{tabular}}
    \caption{A summary of feature embedding-based methods (RGB) regarding year, paradigm and highlight.}
    \label{tab:feat_methods}
\end{table*}

\subsection{Overview of UIAD Methods}

The UIAD that this paper focuses on is essentially a subset of the problems with out-of-distribution (OOD). The goal is to detect abnormal samples in the samples to be tested and to locate the abnormal areas of the abnormal samples. Usually UIAD tasks include anomaly detection and anomaly localization tasks. Traditional anomaly detection uses methods such as machine learning, including clustering methods (such as k-NN \cite{angiulli2002fast}), outlier ensembles (such as isolation forest\cite{liu2008isolation}), and linear models (such as OCSVM \cite{scholkopf2001estimating}, principal component analysis \cite{shyu2003novel}), and others.

With the development of deep learning technology and the arrival of the Industry 4.0, UIAD methods based on deep learning gradually replace traditional methods. Existing UIAD methods based on deep learning mainly use pure RGB images as input, and more and more methods use 3D point clouds or multimodal input to adapt to specific tasks or improve the performance of original 2D tasks. We summarize the current deep learning-based UIAD methods in different modal settings, and summarize the main RGB UIAD methods in this paper in Table \cref{tab:feat_methods,tab:rec_methods,tab:lm_methods} and the main 3D \& multimodal UIAD methods in this paper in Table \ref{tab:methods}. At the same time, we investigated the processes of existing methods in different settings and classified the paradigms of UIAD methods in multiple settings.

\section{RGB UIAD Methods}

The main 2D UIAD methods use RGB images as input. Thus, in this paper, we use RGB UIAD to replace 2D UIAD. As the most mainstream modal setting in current UIAD tasks, RGB UIAD has attracted the most attention, possessing the largest number of datasets and methods. It has also spawned several subtasks aimed at comprehensively enhancing detection accuracy and performance. Currently, there are two mainstream paradigms in RGB UIAD: feature embedding-based methods (as shown in Table \ref{tab:feat_methods}), reconstruction-based methods (as shown in Table \ref{tab:rec_methods}) and large model-based methods (as shown in Table \ref{tab:lm_methods}).

\subsection{Feature Embedding-based Methods}

Feature embedding-based methods use deep learning models pre-trained on large datasets to extract features from samples. These features or representations are then used to locate anomaly areas through additional detection and segmentation submodules. Common architectures in this kind of method include teacher-student architecture, one-class classification, distribution map, and memory bank.

\begin{figure}[ht]
    \centering
    \includegraphics[width=1\columnwidth]{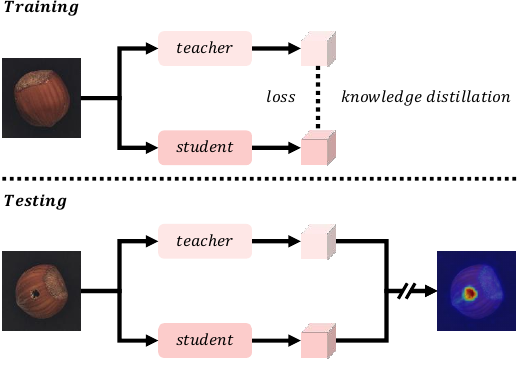}
    \caption{Process of teacher-student architecture paradigm.}
    \label{teacher_student}
\end{figure}

\subsubsection{Teacher-Student Architecture Methods}

The process of teacher-student architecture paradigm is shown in Figure \ref{teacher_student}. US \cite{bergmann2020uninformed} designed a basic framework for solving image anomaly detection tasks using teacher-student architecture. By using a pre-trained teacher network to constrain an untrained student network on the training set containing only normal samples, the student network learns the data distribution of normal samples and outputs similar representations to those of the teacher network on normal samples. During inference phase on the test set, since the student network has not learned the data distribution of abnormal samples, its representation ability on abnormal samples significantly differs from that of the teacher network. MKD \cite{salehi2021multiresolution} trains a clone network (student network) to mimic the behavior of a pre-trained source network (teacher network) on normal data, optimizing with a loss function that combines activation value distance and directional consistency. MKD detects and locates anomalies by comparing the differences between the clone and source networks on the data. GP \cite{wang2021glancing} extracts features from local patches and surrounding areas of images through local and global networks, using inconsistency detection head and distortion detection head to compare differences between local and global features, generating anomaly score maps. PFM \cite{wan2022unsupervised} introduces pre-trained feature mapping (PFM), mapping images from a pre-trained source space (teacher) to a target space (student) while employing bidirectional and multi-hierarchical feature mapping to enhance anomaly detection and segmentation performance. MemKD \cite{gu2023remembering} combines teacher-student architecture with memory bank, designing a normality recall memory (NR Memory) module that alleviates the "normality forgetting" issue of the student network by retrieving features from stored normal information, and utilizes a normality embedding learning strategy to help the NR Memory module learn and remember normality knowledge from normal data. RD4AD \cite{deng2022anomaly} introduced a simple and effective reverse distillation method and eliminated redundant features across multiple scales through a trainable One-Class Bottleneck Embedding module (OCBE). DeSTSeg \cite{zhang2023destseg} uses anomaly simulation and a denoising teacher-student network to make the student network learn more powerful representations. EfficientAD \cite{batzner2024efficientad} proposes a more efficient network based on the teacher-student architecture, achieving real-time inference with millisecond latency. Methods based on the teacher-student architecture discriminate between normal and anomaly areas through the difference in representation ability between the student and teacher networks. EMMFRKD \cite{tong2024enhanced} combines teacher-student architecture and memory bank, utilizing reverse distillation. It designs an enhanced mutual mapping feature fusion module to avoid feature redundancy and introduces a coordinated attention mechanism module to improve anomaly localization accuracy. Additionally, it employs a single-category embedding memory bank to suppress the expression of anomaly information, aiding the student network in reconstructing normal samples. AEKD \cite{wu2024aekd} designs an asymmetric teacher-student network structure to alleviate the generalization problem of student model, using a multi-scale feature fusion module to aggregate multi-layer features. It enhances the feature expression differences between the teacher and student networks through various data flow designs. FCACDL \cite{zhang2024feature} constrains the consistency between the student local view and the teacher global view, maintaining feature centers consistent between the features of the teacher and student, and enhances the learning capability of the student network through consistency constraints in the encoding layer. DMDD \cite{liu2024dual} decouples the teacher-student network to classify features into normality and abnormality, conducting normality guidance modeling and abnormality inverse mimicking distillation separately to ensure significant feature representation differences in normal and anomaly areas for the student network.

\begin{figure}[ht]
    \centering
    \includegraphics[width=1\columnwidth]{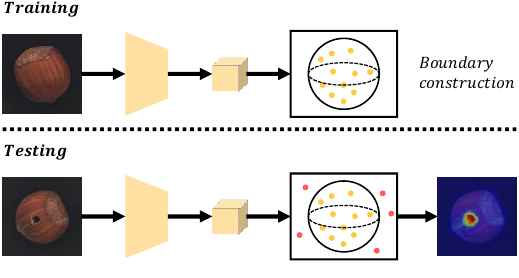}
    \caption{Process of one-class classification paradigm.}
    \label{one-class}
\end{figure}

\subsubsection{One-Class Classification Methods}

The process of one-class classification paradigm is shown in Figure \ref{one-class}. Compared with other methods, one-class classification methods rely more on pseudo samples obtained by anomaly simulation. CutPaste \cite{li2021cutpaste} proposes an anomaly simulation method to generate pseudo samples by cutting small rectangular patches from normal training images and pasting the patches back to images at a random location, and uses CNN to build a self-supervised learning classifier to classify normal samples and augmented samples (pseudo samples). However, there is still a certain gap between the pseudo samples obtained by the anomaly simulation method and the real anomaly samples. CutPaste simulates abnormalities at the image level, while SimpleNet \cite{liu2023simplenet} simulates anomalies in the feature space and uses MLP to build a discriminator to classify normal features and anomaly features. ADShift \cite{cao2023anomaly} minimizes the gap between the source and target distributions during the training phase by proposing generalized normality learning, while during the inference phase, it employs feature distribution matching for data augmentation to improve the generalization performance, thereby enhancing anomaly detection performance under distribution shift conditions. DS2 \cite{he2024learning} refers to two-stage instance-level self-supervised learning (SSL), introducing a two-stage, dense pre-training model for anomaly localization tasks, combining dual positive-pair selection criteria and dual feature scales, and evaluating the learned representations through a generative classifier. GeneralAD \cite{strater2024generalad} designed a self-supervised anomaly feature generation module that creates anomaly features by adding noise and disrupting features during training, and employs a cross-patch attention discriminator to detect anomalies. GLASS \cite{chen2024unified} combines global and local anomaly synthesis strategies to generate pseudo samples, and inputs normal features, global anomaly features and local anomaly features into a discriminator for joint training.

\begin{figure}[ht]
    \centering
    \includegraphics[width=1\columnwidth]{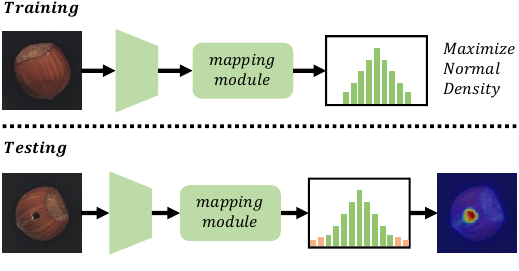}
    \caption{Process of distribution map paradigm.}
    \label{distribution-map}
\end{figure}

\begin{table*}\footnotesize
    \centering
    \renewcommand\arraystretch{1.0}
    \resizebox{\linewidth}{!}{
    \begin{tabular}{lclp{11cm}}
        \toprule
        \textbf{Methods} & \textbf{Year} & \textbf{Paradigm} & \textbf{Highlight} \\
        \midrule
        AE-SSIM \cite{bergmann2018improving} & 2018 & Autoencoder-based & Introduced structural similarity (SSIM) metric as a loss function for autoencoders. \\
        DFR \cite{yang2020dfr} & 2020 & Autoencoder-based & Used autoencoders to reconstruct hierarchical multi-scale regional feature representations. \\
        DAAD \cite{hou2021divide} & 2021 & Autoencoder-based & Introduced multi-scale memory banks to assist with reconstruction. \\
        RIAD \cite{zavrtanik2021reconstruction} & 2021 & Autoencoder-based & Proposed reconstructing images by randomly removing parts and restoring the image. \\
        DRÆM \cite{zavrtanik2021draem} & 2021 & Autoencoder-based & Designed a realistic anomaly simulation strategy using Perlin noise at the image level. \\
        DSR \cite{zavrtanik2022dsr} & 2022 & Autoencoder-based & Simulated anomalies at the feature level instead of the image level. \\
        NSA \cite{schluter2022natural} & 2022 & Autoencoder-based & Introduced seamlessly blending different regions of normal images to generate pseudo samples to optimize model performance. \\
        SSPCAB \cite{ristea2022self} & 2022 & Autoencoder-based & Implemented an anomaly detection plugin using masked convolution and channel attention mechanism. \\
        SSMCTB \cite{madan2023self} & 2022 & Autoencoder-based & Designed a self-supervised masked convolutional transformer block to replace the masked convolution module in SSPCAB. \\
        THFR \cite{guo2023template} & 2023 & Autoencoder-based & Used multi-level memory banks and bottleneck compression to assist with feature-level reconstruction. \\
        FastRecon \cite{fang2023fastrecon} & 2023 & Autoencoder-based & Proposed a training-free, few-shot method that achieve fast feature reconstruction through a linear transformation. \\
        RealNet \cite{zhang2024realnet} & 2024 & Autoencoder-based & Used a diffusion model to generate pseudo samples containing simulated anomalies. \\
        IFgNet \cite{chen2024implicit} & 2024 & Autoencoder-based & Combined the foreground detection task with the reconstruction task to reduce the interference of background noise. \\
        LAMP \cite{park2024neural} & 2024 & Autoencoder-based & Proposed a strategy for loss amplification that does not require changes to the network structure. \\
        PatchAnomaly \cite{fan2024patch} & 2024 & Autoencoder-based & Proposed using patch-level anomalies synthesized through self-supervised learning to provide supervision signals for reconstruction networks. \\
        MAAE \cite{liu2024mixed} & 2024 & Autoencoder-based & Proposed an autoencoder that combines spatial-wise and channel-wise self-attention. \\
        DC-AE \cite{zhang2024dual} & 2024 & Autoencoder-based & Introduced dual constraints of adversarial learning and global memory bank, along with an adaptive weighted similarity spatial attention mechanism. \\
        \midrule
        SCADN \cite{yan2021learning} & 2021 & GAN-based & Introduced multi-scale striped masks and GAN to reconstruct samples. \\
        OCR-GAN \cite{liang2023omni} & 2023 & GAN-based & Used GAN to perform omni-frequency reconstruction of different frequency band information of the sample. \\
        \midrule
        MeTAL \cite{de2022masked} & 2022 & Transformer-based & Introduced the masked self-attention mechanism and multi-resolution patches. \\
        FOD \cite{yao2023focus} & 2023 & Transformer-based & Established intra- and inter-patch correlations to detect and locate anomalies. \\
        AMI-Net \cite{luo2024ami} & 2024 & Transformer-based & Introduced dynamic adaptive masks to effectively obscure anomaly areas. \\
        PNPT \cite{yao2024prior} & 2024 & Transformer-based & Introduced prior information of normal samples as prompts to guide models in reconstructing samples. \\
        \midrule
        DDAD \cite{mousakhan2023anomaly} & 2023 & Diffusion-based & Used a diffusion model as the reconstruction network. \\
        DiffAD \cite{zhang2023unsupervised} & 2023 & Diffusion-based & Combined feature-level diffusion models and channel interpolation to address the misalignment problem. \\
        RAN \cite{lu2023removing} & 2023 & Diffusion-based & Introduced multi-scale noise and used KL divergence and MSE to calculate pixel-level and feature-level anomaly scores, respectively. \\
        TransFusion \cite{fuvcka2023transfusion} & 2024 & Diffusion-based & Introduced a transparency-based diffusion model to solve the problems of overgeneralization and loss of detail. \\
        DiAD \cite{he2024diffusion} & 2024 & Diffusion-based & Combined semantic guidance with diffusion models to avoid category misclassification and semantic loss during reconstruction. \\
        GLAD \cite{yao2024glad} & 2024 & Diffusion-based & Proposed a diffusion model that combines global and local adaptive mechanisms to improve the reconstruction performance for large-scale anomalies. \\
        AnomalySD \cite{yan2024anomalysd} & 2024 & Diffusion-based & Combined a stable diffusion model with a multi-level masking approach to focus on reconstructing the anomaly areas. \\
        \bottomrule
    \end{tabular}}
    \caption{A summary of reconstruction-based methods (RGB) regarding year, paradigm and highlight.}
    \label{tab:rec_methods}
\end{table*}

\subsubsection{Distribution Map Methods}

The process of distribution map paradigm is shown in Figure \ref{distribution-map}. Among distribution map methods, normalized flow is the most mainstream method. DifferNet \cite{rudolph2021same} introduces normalizing flow (NF) into UIAD tasks, using NF for density estimation, assigning a likelihood value to each image, and calculating anomaly score based on the likelihood value. FastFlow \cite{yu2021fastflow} simulates global and local distribution by alternately stacking large and small convolution kernels, and achieves efficient and lightweight end-to-end inference. CS-Flow \cite{rudolph2022fully} uses images of different sizes as input, inputs multi-scale features into the normalized flow network, and obtains the anomaly score through cross-scale density estimation. CFlow-AD \cite{gudovskiy2022cflow} introduces position encoder into the conditional normalized flow network and obtains the anomaly score through the estimated multi-scale likelihoods. CDO \cite{cao2023collaborative} adds random perturbations to input samples to generate pseudo samples. The discrepancy distribution is obtained through the discrepancy between teacher and student network outputs. It minimizes the difference between normal features while maximizing the difference between anomaly features, and further optimizes the tail samples to reduce the overlap between normal and anomaly distributions. PyramidFlow \cite{lei2023pyramidflow} introduces a latent template-based defect contrastive localization paradigm, performs contrast localization in latent space, effectively reduces the intra-classes variance, and proposes invertible pyramids and pyramid coupling blocks for multi-scale fusion and mapping, achieving high-resolution anomaly detection and localization. SLAD \cite{xu2023fascinating} introduces scale learning, using data sub-vector transformed representations and labels generated from sub-vector scales to drive neural network training. SLAD optimizes the difference between the predicted distribution of network and the target distribution, allowing the model to better describe the normality of normal samples. MSFlow \cite{zhou2024msflow} uses asymmetrical parallel flows followed by a fusion flow to exchange multi-scale perceptions, and uses different multi-scale aggregation strategies to detect and locate anomalies. AttentDifferNet \cite{simoes2024attention} uses attention modules based on normalizing flow for density estimation, allowing the model to focus on foreground objects and optimizing its attention to key features.

\begin{figure}[ht]
    \centering
    \includegraphics[width=1\columnwidth]{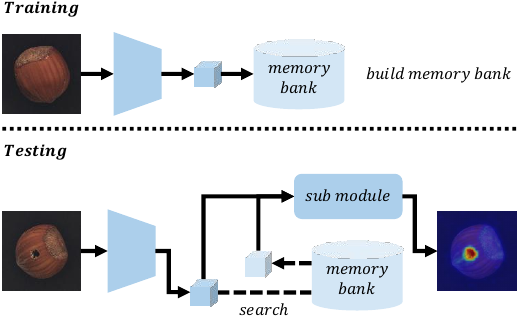}
    \caption{Process of memory bank paradigm.}
    \label{memory-bank}
\end{figure}

\subsubsection{Memory Bank Methods}

The process of memory bank paradigm is shown in Figure \ref{memory-bank}. Memory bank methods rely on a pre-trained feature extractor to extract image features and requires additional memory space to store normal image features. In the inference phase, anomalies are detected and located by comparing the input image features with the features in the memory bank. PaDiM \cite{defard2021padim} extracts patch embedding vectors from normal images using a pre-trained convolutional neural network and obtains a probabilistic representation of normal samples through a multivariate Gaussian distribution. Anomaly scores are derived by calculating the Mahalanobis distance between the patch embeddings of test images and the Gaussian distribution saved during the training phase. PatchCore \cite{roth2022towards} uses a maximally representative memory bank of normal patch-features, and introduces a greedy coreset subsampling method to reduce the redundancy of the memory bank to reduce storage memory and inference time. CFA \cite{lee2022cfa} clusters normal features by constructing coupled hyperspheres, adopts transfer learning on the target dataset to alleviate the bias of pre-trained CNNs, and reduces inference cost through adaptively compressed memory banks. DMAD \cite{liu2023diversity} employs VQ-Layer as an information compression module, quantizing the query vector to a single-memory feature cube using L2 distance. DMAD avoids the undesired reconstruction of normal-like anomalies caused by a linear combination of multiple memory items in existing methods, resulting in more compact feature representations. It also estimates multi-scale deformation fields between input images and reconstructed images, distinguishing between anomaly and normal samples based on the degree of deformation. PNI \cite{bae2023pni} uses position information and neighborhood information to more accurately estimate the distribution of normal features in the memory banks. GraphCore \cite{xie2023pushing} utilizes graph neural networks to extract vision isometric invariant features, enabling the construction of a more compact memory bank that remains effective in few-shot settings. InReaCh \cite{mcintosh2023inter} identifies similar image regions across multiple training images, generating channels that span across these images to extract high-confidence normal image patches from training data. It prunes the generated channels to remove those with low span or high spread, detecting anomaly patches by comparing test image patches with those in the normal channels. ReconFA \cite{zuo2024reconstruction} aggregates multi-scale features from normal samples to obtain a more compact multi-scale feature representation. It simultaneously trains an encoder to adapt the extracted features to target domain while reducing the feature dimensions, thereby constructing more compact memory banks with lower spatial complexity. ReConPatch \cite{hyun2024reconpatch} trains a simple linear transformation through pairwise similarity and contextual similarity to build target-oriented, easily distinguishable feature representations in the memory banks.

\begin{figure}[ht]
    \centering
    \includegraphics[width=1\columnwidth]{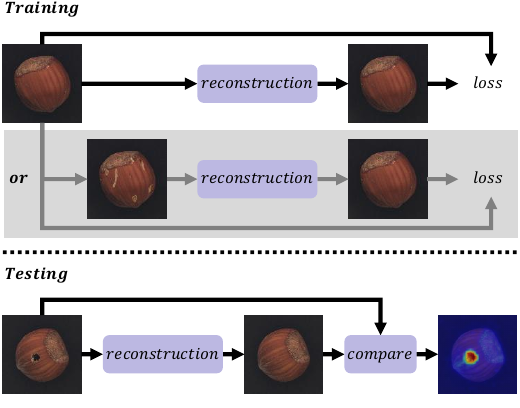}
    \caption{Process of reconstruction paradigm.}
    \label{reconstruction}
\end{figure}

\subsection{Reconstruction-based Methods}

Reconstruction-based methods mainly use reconstruction networks as the main module, as shown in Figure \ref{reconstruction}. The reconstruction network learns the distribution of normal samples to reconstruct the input samples to be detected, and detects and locates anomalies through reconstruction errors. The reconstruction error is based on the following assumptions: normal areas can be accurately reconstructed by the model, while abnormal areas are difficult to reconstruct. The reconstruction network has a low degree of dependence on pre-trained models and usually learns feature representations autonomously. Reconstruction-based methods locate anomalies by comparing pixel-level reconstruction errors, so they are superior to feature embedding-based methods in terms of pixel-level indicators. Current reconstruction-based RGB UIAD methods mainly use autoencoders, GANs, Transformers, and diffusion models as reconstruction networks.

\subsubsection{Autoencoder-based Methods}

Autoencoders are currently the most widely used reconstruction networks, characterized by a relatively simple structure comprising an encoder and a decoder. The encoder compresses input images into low-dimensional latent space representations, while the decoder reconstructs original images from these representations. The encoder-decoder architecture is easy to implement and has low computational costs. When computing reconstruction errors, traditional autoencoders typically use pixel-based loss functions (such as L2 distance) for reconstruction. This approach can lead to larger residuals at edge locations and fails to effectively detect structural anomalies with roughly the same pixel intensity. AE-SSIM \cite{bergmann2018improving} applies structural similarity (SSIM) metric as a loss function to autoencoders to address the shortcomings of traditional pixel-based reconstruction error methods. DFR \cite{yang2020dfr} uses a pre-trained CNN to extract hierarchical multi-scale regional feature representations from input images and employs an autoencoder to reconstruct these features, detecting and locating anomalies by comparing the features before and after reconstruction. DAAD \cite{hou2021divide} proposed that finer granularity leads to better reconstruction, while coarser granularity leads to poorer reconstruction. DAAD divided feature maps into blocks of varying sizes and used a memory bank for each block to store normal patterns, preventing excessive reconstruction of anomaly samples. Additionally, it introduced adversarial learning to detect and locate anomalies from images before and after reconstruction using a discriminator. RIAD \cite{zavrtanik2021reconstruction} proposes randomly removing parts of the image areas to reduce the ability of models to accurately reconstruct anomalies. It uses an autoencoder to restore images at multiple scales and generates anomaly scores by comparing the structured similarity and multi-scale gradient magnitude similarity between the original and restored images. DRÆM \cite{zavrtanik2021draem} employs a more realistic anomaly simulation strategy, generating anomaly masks using Perlin noise and using external datasets for filling textures. DRÆM generates pseudo samples that can be regarded as abnormal samples at the image level, and then uses an autoencoder as reconstruction network, and designs a segmentation network to learn the differences between pseudo samples and reconstruction results to detect and locate anomalies. Different from DRÆM, DSR \cite{zavrtanik2022dsr} simulates anomalies at the feature level to generate pseudo samples, uses an autoencoder as reconstruction network, and expands it into a dual-branch structure with discrete latent representations. NSA \cite{schluter2022natural} proposed using Poisson image editing techniques to seamlessly blend different areas of normal images, generating synthetic anomaly samples with a natural appearance. These synthetic samples are then used to enhance the ability of models to detect anomalies. SSPCAB \cite{ristea2022self} proposed a "plug-and-play" module that masks the central area of the convolutional kernel, forcing the model to reconstruct the masked area using contextual information. It then recalibrates the feature maps from the convolution operation via the channel attention mechanism, highlighting important features and suppressing irrelevant features to detect anomalies. SSMCTB \cite{madan2023self} extends the SSPCAB by replacing standard channel attention with multi-head self-attention and expanding masked convolution into 3D convolutional filters. Additionally, the mean squared error (MSE) loss is replaced with the Huber loss, which is less sensitive to outliers. THFR \cite{guo2023template} extracts image-level features from normal samples and stores them as template features in memory banks. It then hierarchically compresses image features through global and local bottleneck structures while using the memory banks for hierarchical compensation of the compressed features to restore them to normal features. The anomaly score is calculated based on the difference between the features before and after restoration. FastRecon \cite{fang2023fastrecon} proposed an efficient reconstruction method suitable for few-shot scenarios. It constructs support sample feature banks using patch-level features from a small number of normal samples. The sample to be detected is treated as a query sample, and the optimal transformation between the support sample features and the query sample features is estimated through a regression algorithm with distribution regularization. This ensures that the reconstructed sample and the query sample exhibit similarity in normal areas. The reconstructed features are compared with the original sample to obtain the anomaly score. RealNet \cite{zhang2024realnet} uses a diffusion model to generate pseudo samples containing simulated anomalies, which are closer to real anomaly samples than previous methods, and filters pre-trained features before reconstruction and uses an autoencoder as the reconstruction network. IFgNet \cite{chen2024implicit} combines multi-task learning by designing sub-networks that simultaneously perform foreground detection and image reconstruction tasks. It extracts shared features through a common network and leverages the results of foreground detection to optimize the accuracy of anomaly detection, thereby reducing the interference of background noise. Based on the baseline using autoencoders for reconstruction, LAMP \cite{park2024neural} modifies the shape of the loss function by amplifying the reconstruction loss, making the reconstruction error curve steeper. This amplifies the reconstruction error of anomaly samples, limiting the generalization ability of models to anomaly samples without requiring additional changes to the network structure. PatchAnomaly \cite{fan2024patch} utilizes a patch-level self-supervised data augmentation method to generate pseudo samples that closely resemble real anomaly samples, providing reliable supervision signals for reconstruction networks. The features before and after reconstruction are then fed into a subsequent detection head to calculate anomaly scores. MAAE \cite{liu2024mixed} generates adaptive noise to perturb features based on different object categories, creating simulated anomalies. It captures the global shapes of objects through spatial-wise self-attention, while channel-wise self-attention alleviates the color distortions caused by various lighting conditions. By integrating the low-level and the high-level features, MAAE continuously adjust the low-level features as the higher-level features are added to alleviate the semantic blur problem and appropriately preserve the surface semantics of the subtle anomalies. DC-AE \cite{zhang2024dual} introduces dual constraints of adversarial learning and global memory bank to autoencoders, suppressing excessive reconstruction of anomaly features and preventing over-restoration of anomaly areas. At the same time, by calculating the cosine similarity between normal and anomaly patterns, it enhances the features of anomaly areas while suppressing interference from background areas.

\subsubsection{GAN-based Methods}

GANs (Generative Adversarial Networks) \cite{goodfellow2014generative} consist of a generator and a discriminator. Through adversarial training, the generator learns the data distribution to create realistic images. While GANs are better at capturing the complex distributions and features of data, they are less stable than autoencoders and are prone to mode collapse. SCADN \cite{yan2021learning} randomly removes a part of regions in images using multi-scale striped masks, using a GAN to learn the semantic features of images based on the surrounding contextual information. During the training phase, it minimizes both the difference between the original and reconstructed samples by the generator and the adversarial loss. In the testing phase, the generator reconstructs the removed regions, and the anomaly score map is obtained by comparing the images before and after reconstruction. OCR-GAN \cite{liang2023omni} extracts different frequency band information of the sample to be detected, uses GAN to enable the omni-frequency reconstruction by multiple branches, and realizes the interaction between different frequency band information to enhance reconstruction.

\subsubsection{Transformer-based Methods}

Transformers \cite{vaswani2017attention}, based on self-attention mechanism, can capture global features and long-distance relationships while simultaneously attending to local image details and overall structure. However, the computational complexity of the self-attention mechanism increases quadratically with the input length (up to $O(N^2)$), affecting training efficiency. MeTAL \cite{de2022masked} proposes a new model based on the Vision Transformer \cite{dosovitskiy2020image} architecture with patch masking. When reconstructing images, it masks the information of each image patch and relies solely on the information from neighboring patches to reconstruct the masked patch, incorporating multi-resolution patches in the process. FOD \cite{yao2023focus} uses the modeling ability of Transformers to perform feature-level reconstruction of input samples at the feature level, and establishes intra- and inter-patch correlations. FOD detects and localizes anomalies by correlating differences and reconstruction errors. AMI-Net \cite{luo2024ami} proposes the use of dynamic adaptive masks to effectively obscure anomaly-related features, replacing traditional fixed or random masks. During the training phase, AMI-Net randomly masks different positions of images, enabling the model to learn to restore normal areas under various anomaly sizes and locations. In the testing phase, it dynamically generates masks to obscure anomaly-related areas and repairs them using global contextual information, avoiding the possibility of reconstructing the anomaly areas. PNPT \cite{yao2024prior} introduces the concept of " prior normality prompt", extracting and storing the features of normal samples for each category to provide prior information about normality. This prior information is then used as prompts to guide models in reconstructing anomaly samples, thereby avoiding the occurrence of the "identical mapping" problem.

\subsubsection{Diffusion-based Methods}

Diffusion models define a forward diffusion process that progressively adds noise and a reverse denoising process that learns to remove the noise, gradually generating reconstructed images. Due to high-fidelity generation capabilities, diffusion models have become one of the preferred choices for reconstruction networks. However, diffusion models are complex to implement and require multiple iterative steps, making the training and inference processes time-consuming and computationally expensive, rendering them unsuitable for certain scenarios. DDAD \cite{mousakhan2023anomaly} uses a diffusion model as the reconstruction network, fine-tunes a pre-trained feature extractor through domain adaptation, and locates anomalies based on pixel-level errors and feature-level errors before and after reconstruction. DiffAD \cite{zhang2023unsupervised} simulates the generation of pseudo-anomalous samples on the input data and encodes both the input and pseudo-samples into latent vectors, reconstructing them using a feature-level diffusion model. The pseudo-sample features and reconstructed features are interpolated and decoded to generate interpolated samples. These additional interpolated samples are concatenated with the pseudo-samples and reconstructed samples along the channel dimension to mitigate the anomaly localization error caused by subtle differences between the normal pixels of the reconstructed and pseudo-samples. RAN \cite{lu2023removing} introduces multi-scale noise into input images and reconstructs the noisy images using diffusion models. KL divergence is used to calculate pixel-level differences before and after reconstruction, while MSE is used to compute feature-level differences. TransFusion \cite{fuvcka2023transfusion} identifies two challenges faced by reconstruction models: overgeneralization and loss of detail. To address these issues, TransFusion proposes a transparency-based diffusion process, where the transparency of the anomaly areas gradually increases, eventually replacing them with their corresponding normal appearance. By performing reconstruction and localization simultaneously, TransFusion is able to gradually erase anomalies while preserving the integrity of normal areas. DiAD \cite{he2024diffusion} utilizes a diffusion model for multi-class anomaly detection and designs a semantic-guided network, stably connected to the denoising network of the diffusion model. This approach preserves the semantic information of the original image while reconstructing anomalous regions, preventing category misclassification and semantic loss during reconstruction. Anomaly score maps are calculated based on multi-scale features before and after reconstruction. GLAD \cite{yao2024glad} adaptively determines the denoising steps based on the image content and anomaly type, avoiding excessive denoising that could impact the details of normal areas. It also introduces synthetic anomaly samples during training, breaking the limitation of the standard Gaussian distribution in diffusion models. During the inference phase, GLAD integrates features from normal areas with reconstructed features from anomaly areas, ensuring that the details of normal areas are preserved while suppressing the reconstruction of anomaly areas. AnomalySD \cite{yan2024anomalysd} fine-tunes a stable diffusion denoising network to reconstruct only few shots of normal samples that are partially masked. During the inference phase, a dual-stream reconstruction strategy is adopted. First, a multi-scale patch-by-patch masking approach is used, followed by reconstruction through the fine-tuned stable diffusion model. Then, error maps of input samples, computed from a normal prototype bank, are used to add masks, with reconstruction again performed through the fine-tuned stable diffusion model. Finally, the results of two streams are fused to obtain final anomaly score maps.

\begin{table*}
    \centering
    \renewcommand\arraystretch{1.0}
    \resizebox{\linewidth}{!}{
    \begin{tabular}{lclp{11cm}}
        \toprule
        \textbf{Methods} & \textbf{Year} & \textbf{Paradigm} & \textbf{Highlight} \\
        \midrule
        AnomalyGPT \cite{gu2024anomalygpt} & 2023 & GPT-based & Applied LVLMs to industrial anomaly detection tasks for the first time. \\
        Myriad \cite{li2023myriad} & 2023 & GPT-based & Integrated the visual expert model to convert prior knowledge into features which are able to be understood by LLM. \\
        GPT-4V-AD \cite{zhang2023exploring} & 2023 & GPT-based & Applied the visual question answering paradigm to zero-shot anomaly detection tasks. \\
        Customizable-VLM \cite{xu2024customizing} & 2024 & GPT-based & Customized a generic visual-language model using a multimodal prompting strategy. \\
        LogiCode \cite{zhang2024logicode} & 2024 & GPT-based & Combined the logical reasoning capabilities of LLMs to automatically generate Python codes for detecting and explaining logical anomalies. \\
        \midrule
        WinCLIP \cite{jeong2023winclip} & 2023 & CLIP-based & Introduced CLIP for the first time to achieve zero-/few-shot UIAD. \\
        ClipSAM \cite{you2022unified} & 2023 & CLIP-based & Interacted language features with visual features at both row-column and multi-scale levels. \\
        CLIP-AD \cite{chen2023clip} & 2023 & CLIP-based & Introduced CLIP Surgery to reduce redundant features, aiming to alleviate opposite predictions and irrelevant highlights. \\
        AnoCLIP \cite{deng2023anovl} & 2023 & CLIP-based & Focused on extracting local visual tokens from the CLIP visual encoder. \\
        AnomalyCLIP \cite{zhou2023anomalyclip} & 2024 & CLIP-based & Introduced an object-agnostic prompt learning method to improve the cross-domain generalization ability of zero-shot UIAD. \\
        CLIP-ADA \cite{cai2024anomaly} & 2024 & CLIP-based & Proposed learnable prompts to enhance the generalization capability of CLIP in UIAD. \\
        PromptAD \cite{li2024promptad} & 2024 & CLIP-based & Constructed negative samples through semantic concatenation and introduced an explicit anomaly margin to optimize prompt learning in one-class anomaly detection. \\
        FiLo \cite{gu2024filo} & 2024 & CLIP-based & Generated fine-grained anomaly descriptions for each category. \\
        DIE-CLIP \cite{zhang2024dualimage} & 2024 & CLIP-based & Proposed inputting a pair of images during the testing phase to provide mutual reference information. \\
        AnoPLe \cite{lee2024anople} & 2024 & CLIP-based & Proposed a bi-directional modal information interaction to optimize prompt learning. \\
        \midrule
        SAA/SAA+ \cite{cao2023segment} & 2023 & SAM-based & Applied SAM and GroundingDINO to UIAD tasks. \\
        UCAD \cite{liu2024unsupervised} & 2024 & SAM-based & Used SAM to partition the training samples based on their structures to make the features in the memory bank more compact. \\
        STLM \cite{li2024sam} & 2024 & SAM-based & Used SAM as a teacher network to distill knowledge to student networks. \\
        \bottomrule
    \end{tabular}}
    \caption{A summary of large model-based methods (RGB) regarding year, paradigm and highlight.}
    \label{tab:lm_methods}
\end{table*}

\subsection{Large Model-based Methods}

Large models usually refer to artificial intelligence models based on deep learning and trained on large-scale datasets. In the field of computer vision, examples include CLIP \cite{radford2021learning}, SAM \cite{kirillov2023segment}, GPT-4V \cite{yang2023dawn} and GPT-4o \cite{openai2024gpt4o}, etc. These models have a large number of parameters and perform well in various visual tasks. However, existing large models lack expert knowledge in industrial anomaly detection, and training specialized large models requires extensive labeled data and computational resources. Current UIAD methods based on large models primarily solves these issues. We categorize large model-based methods into CLIP-based methods, SAM-based methods and GPT-based methods, summarized in Table \ref{tab:lm_methods}.

\begin{figure}[ht]
    \centering
    \includegraphics[width=1\columnwidth]{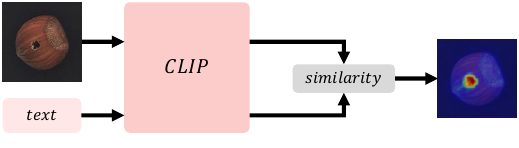}
    \caption{Process of CLIP-based methods.}
    \label{clip}
\end{figure}

\subsubsection{CLIP-based Methods}

CLIP (Contrastive Language-Image Pretraining) \cite{radford2021learning} is a multimodal model that comprehends and processes both images and text. By employing contrastive learning, it maps images and text into a shared embedding space, enabling them to be interrelated. CLIP has demonstrated outstanding performance in many tasks, such as image classification, image segmentation, and zero-shot learning. The process of the CLIP-based UIAD methods is shown in Figure \ref{clip}. WinCLIP \cite{jeong2023winclip} introduces CLIP into UIAD by encoding input text and images to be detected, aggregating multi-scale image features and text features to ensure alignment between vision and language. It introduces a compositional prompt ensemble method that combines predefined state words with various text templates to generate text embeddings, better expressing object states and enhancing the understanding ability of CLIP models in zero-/few-shot anomaly detection tasks. ClipSAM \cite{you2022unified} enhances the interaction between language and visual features at both row-column and multi-scale levels. This approach assists CLIP in identifying local anomalies from multiple directions and generates more precise anomaly segmentation results using SAM \cite{kirillov2023segment}. CLIP-AD \cite{chen2023clip} notes that directly applying CLIP to anomaly detection can result in opposite predictions and irrelevant highlights. To address these issues, CLIP Surgery \cite{li2023clip} is introduced to reduce invalid predictions caused by redundant features and employs multi-level feature fusion to further improve these problems. Additionally, CLIP-AD uses linear layers on the image features output from each layer of the backbone network, mapping these features into an embedding space aligned with text features to obtain an extended version of the model. However, CLIP focuses on global image-text pairing, which results in an inability to accurately identify local anomalies in fine-grained anomaly localization. AnoCLIP \cite{deng2023anovl} addresses this issue by extracting locally aware visual tokens from CLIP visual encoder and designs a training-free value-to-value attention mechanism to compute these local-aware patch tokens. Additionally, to enhance the fine-grained alignment between vision and language features, AnoCLIP generates rich prompts for both normal and anomaly states. AnomalyCLIP \cite{zhou2023anomalyclip} introduces an object-agnostic prompt learning method that enhances the generalization ability of the model across different domains by learning universal normal and anomaly prompts. This reduces reliance on manually defined prompts and extends applicability to highly diverse data in industrial and medical fields. CLIP-ADA \cite{cai2024anomaly} replaces traditional, manually designed text prompts with learnable prompts and employs self-supervised learning to adaptively associate them with anomaly patterns. This approach enables models to better adapt to anomaly detection tasks across various categories and scenarios. Additionally, by utilizing a coarse-to-fine anomaly region refinement strategy, it treats coarse anomaly localization results as attention maps to further focus and obtain more refined anomaly localization outcomes. PromptAD \cite{li2024promptad} points out that existing many-class prompt learning are unsuitable for one-class anomaly detection. To address this, it proposes semantic concatenation as a novel prompt learning approach. By concatenating normal prompts with anomaly suffixes, it transforms normal prompts into anomaly prompts, providing models with negative samples for contrastive learning. PromptAD introduces an explicit anomaly margin to control the distance between normal and anomaly prompt features, ensuring the distinguishability of normal and anomaly samples in the feature space. FiLo \cite{gu2024filo} employs LLMs to generate fine-grained anomaly descriptions for each category, replacing the previous generic descriptions. It also designs learnable text templates, enabling these detailed descriptions to better match specific anomaly samples. DIE-CLIP \cite{zhang2024dualimage} introduces a dual-image enhancement strategy that inputs a pair of unlabeled images during the testing phase. Each image serves as a visual reference for the other, and this mutual information is integrated into language-visual predictions of CLIP to enhance the robustness of anomaly detection. AnoPLe \cite{lee2024anople} finds that relying on single-modal (image or text) prompt learning is insufficient when anomaly samples are absent. To address this issue, it proposes a bi-directional prompt learning method that introduces learnable prompts into both visual and textual encoders. By linear projecting the prompts of each modal to the hidden dimensions of the other, this approach enables information exchange between the two modals, thereby reducing dependence on anomaly prior knowledge.

\begin{figure}[ht]
    \centering
    \includegraphics[width=1\columnwidth]{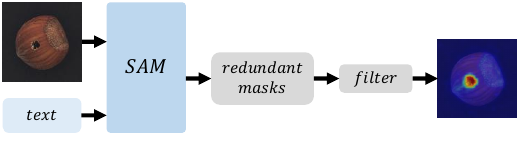}
    \caption{Process of SAM-based methods.}
    \label{sam}
\end{figure}

\subsubsection{SAM-based Methods}

SAM (Segment Anything Model) \cite{kirillov2023segment} is a universal image segmentation model trained on the world's largest segmentation dataset, SA-1B (includes more than 1 billion masks from 11 million licensed and privacy-preserving images). It can generate precise segmentation masks for any object based on minimal input, possessing strong zero-shot learning capabilities. Without additional training, it generalizes to new objects and images, making it highly adaptable across different domains. The process of the SAM-based UIAD methods is shown in Figure \ref{sam}. SAA \cite{cao2023segment} inputs the image to be detected and a language prompt into GroundingDINO \cite{liu2023grounding}, roughly retrieving coarse anomaly region proposals (bounding-box-level regions and their corresponding confidence scores). These proposals are then refined into high-quality, pixel-level anomaly segmentation maps using SAM. Additionally, as an extended version of SAA, SAA+ introduces finer-grained anomaly descriptions as prompts and extracts saliency and region confidence from the image to enhance the reliable prediction of anomaly regions. UCAD \cite{liu2024unsupervised} does not directly use SAM to generate anomaly segmentation results. Instead, it leverages the powerful segmentation capabilities of SAM to produce segmentation maps, where different areas represent distinct structures. By drawing features of areas with the same structure closer together and pushing those of different structures further apart, it achieves a more compact normal feature knowledge base. UCAD calculates anomaly scores by comparing the features of the image to be detected with those in the knowledge base. STLM \cite{li2024sam} also does not use SAM to directly generate anomaly segmentation maps, instead, it leverages SAM as a teacher network to distill knowledge to student network. STLM employs a dual student stream architecture: plain student stream and denoising student stream. The plain student stream is trained to produce discriminative and general feature representations in both normal and anomaly regions, while the denoising student stream is trained to ignore anomaly regions and reconstruct the anomaly regions as normal.

\begin{figure}[ht]
    \centering
    \includegraphics[width=1\columnwidth]{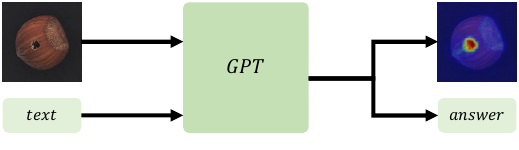}
    \caption{Process of GPT-based methods.}
    \label{gpt}
\end{figure}

\subsubsection{GPT-based Methods}

GPT (Generative Pre-trained Transformer) is a neural network model that uses the transformer architecture. After pre-training on large-scale datasets, it accepts inputs like natural language or images and generates text or image content in a conversational manner. Fine-tuned GPT models can be used for tasks such as dialogue, Q\&A, and content generation. The process of the GPT-based UIAD methods is shown in Figure \ref{gpt}. AnomalyGPT \cite{gu2024anomalygpt} is based on Large Visual Language Models (LVLMs), simulating anomalies on normal samples to generate corresponding anomaly textual descriptions. It designs a lightweight, visual-textual feature-matching-based decoder, directly comparing local visual features with textual descriptions to produce pixel-level anomaly localization results. AnomalyGPT fine-tunes the LVLM by embedding industrial anomaly detection knowledge through a prompt learner, generating prompt embeddings that help the LVLM utilize image inputs, anomaly localization results, and user textual inputs for anomaly detection and localization. Myriad \cite{li2023myriad} selects MiniGPT-4 as the base large multimodal model (LLM), employs a pre-trained zero-shot industrial anomaly detection model as a vision expert, and transforms the prior knowledge provided by the visual expert into textual tokens which are able to be understood by LLM. It then generates vision-language representations aligned with the industrial anomaly detection domain based on the vision experts priors, using a simple domain adapter to bridge the gap between general visual representations and industrial images. Finally, Myriad inputs the textual tokens, vision-language representations, and a text instruction into the LLM to obtain anomaly detection results and further detailed descriptions. GPT-4V-AD \cite{zhang2023exploring} utilizes a framework based on the visual question answering (VQA) paradigm. It begins by employing a super-pixel method to partition the input image into regions. A general prompt description is then designed for all categories, incorporating replaceable category information and region partitioning details. The image with partitioned regions and prompt descriptions are input into GPT-4V. Finally, by combining the structured outputs generated by VQA with the preprocessed region partitioning information, the final anomaly detection results are obtained. Customizable-VLM \cite{xu2024customizing} unifies different types of input data (such as RGB images, point clouds, and time-series data) into a standardized 2D image format, achieving modal unification. It proposes a multimodal prompting strategy that integrates expert domain knowledge, including four types: task instructions, class contexts, normality criteria, and reference normal images, to provide domain-specific knowledge as external memory. Finally, a generic visual-language foundation model is used to obtain binary detection results by utilizing the samples to be detected and the prompts (images or text). LogiCode \cite{zhang2024logicode} detects logical anomalies by automatically generating Python codes using LLMs. It employs a code prompt module to analyze normal and anomaly images, define logical rules, and generate prompts for analysis tasks. Then, a code generation module uses LLMs to convert these logical rules into executable Python codes. Finally, a code execution module runs the generated codes to perform logical and visual analysis of the images, detect and report anomalies, and provide rule-based explanations, thereby simulating the human decision-making process.

\begin{table*}
    \centering
    \renewcommand\arraystretch{1.0}
    \resizebox{\linewidth}{!}{
    \begin{tabular}{clclp{11cm}}
        \toprule
        & \textbf{Methods} & \textbf{Year} & \textbf{Paradigm} & \textbf{Highlight} \\
        \midrule
        \multirow{7}*{\rotatebox{90}{3D}}
        & 3D-ST \cite{bergmann2023anomaly} & 2023 & Teacher-student architecture & Designed a 3D student teacher network for 3D point cloud data. \\
        & Reg3D-AD \cite{liu2023real3d} & 2024 & Memory bank & Introduced memory banks for the first time in 3D point cloud anomaly detection. \\
        & Group3AD \cite{zhu2024towards} & 2024 & Memory bank & Expressed the structural information of point clouds through group-level features. \\
        & PointCore \cite{zhao2024pointcore} & 2024 & Memory bank & Used a single memory bank to store both local and global features to reduced computational complexity. \\
        & R3D-AD \cite{zhou2024r3d} & 2024 & Reconstruction-based & Designed an anomaly simulation strategy for 3D point clouds and used a diffusion model to reconstruct pseudo-samples of 3D point clouds. \\
        \midrule
        \multirow{22}*{\rotatebox{90}{Multimodal}}
        & PatchCore+FPFH \cite{horwitz2023back} & 2023 & - & Designed an feature description method based on point clouds. \\
        & AST \cite{rudolph2023asymmetric} & 2023 & Teacher-student architecture & Designed an asymmetric student-teacher network architecture. \\
        & MMRD \cite{gu2024rethinking} & 2024 & Teacher-student architecture & Proposed a multimodal reverse distillation method and achieved parameter-free fusion of different modal features. \\
        & M3DM \cite{wang2023multimodal} & 2023 & Memory bank & Introduced multiple memory banks to multimodal anomaly detection for the first time and using contrastive learning for multimodal feature fusion. \\
        & CPMF \cite{cao2024complementary} & 2023 & Memory bank & Used a single memory bank to store the aggregated multimodal fusion features. \\
        & Shape-Guided \cite{chu2023shape} & 2023 & Memory bank & Used neural implicit functions of signed distance fields to represent local shapes. \\
        & LSFA \cite{tu2024self} & 2024 & Memory bank & Built global-level and local-level dynamically updated memory banks. \\
        & ITNM \cite{wang2024incremental} & 2024 & Memory bank & Adopted an incremental training method to continuously update the memory bank with new samples. \\
        & CMDIAD \cite{sui2024cross} & 2024 & Memory bank & Proposed a Multi-modal Training, Few-modal Inference (MTFI) pipeline to address the issue of modal missing. \\
        & M3DM-NR \cite{wang2024m3dm} & 2024 & Memory bank & Designed a two-stage denoising network to denoise the input training data to alleviate data noise issues in real-world environments. \\
        & EasyNet \cite{chen2023easynet} & 2023 & Reconstruction-based & Reconstructed the input samples without pre-trained feature extractors and memory banks and proposed an information entropy-based multimodal feature fusion strategy. \\
        & DBRN \cite{bi2023dual} & 2023 & Reconstruction-based & Proposed an importance scoring module to fuse multimodal features. \\
        & 3DSR \cite{zavrtanik2024cheating} & 2024 & Reconstruction-based & Proposed a new depth map anomaly simulation strategy using a randomized affine transform. \\
        & CFM \cite{costanzino2023multimodal} & 2024 & Reconstruction-based & Proposed a lightweight cross-modal mapping network as a reconstruction network. \\
        & 3DRÆM \cite{zavrtanik2024keep} & 2024 & Reconstruction-based & Extend DRÆM to multimodal settings. \\
        \bottomrule
    \end{tabular}}
    \caption{A summary of 3D and multimodal methods regarding year, paradigm and highlight.}
    \label{tab:methods}
\end{table*}

\section{3D UIAD Methods}

\begin{figure}[t]
    \centering
    \includegraphics[width=1\columnwidth]{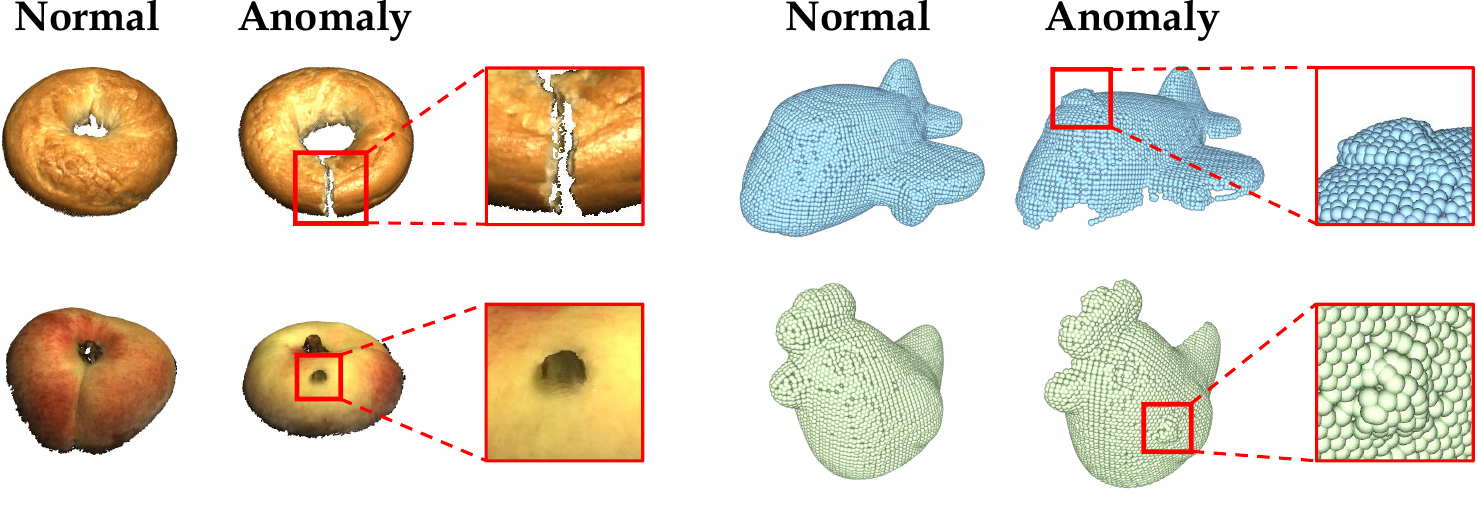}
    \caption{Examples of normal and anomaly samples in multimodal and 3D datasets (selected from MVTec 3D-AD dataset and Real3D-AD dataset).}
    \label{3d-example}
\end{figure}

3D UIAD methods mainly use 3D point clouds or depth maps as inputs, and detect and locate anomaly areas through spatial structure, shape information or depth information. Unlike RGB image inputs, 3D information does not include color information, which results in lower accuracy when detecting anomalies without distinct structural features such as stains and rust spots. However, the accuracy is higher for anomalies with distinct structural features such as cracks, holes, and misalignments. Examples of 3D samples are shown in Fig. \ref{3d-example}. Currently, there are few existing methods for pure 3D UIAD based on 3D point clouds or depth maps.

3D-ST \cite{bergmann2023anomaly} designs a 3D student teacher network for 3D point cloud data, and pre-trained the teacher network through local geometric descriptors and additional data. Local geometric descriptors are obtained from the input point cloud through a local feature extractor. Reg3D-AD \cite{liu2023real3d} is the first to introduce memory banks into 3D point cloud anomaly detection, addressing the lack of baseline methods for high-resolution 3D point cloud anomaly detection. Group3AD \cite{zhu2024towards} expresses the structural information of point clouds through group-level features and constructs contrastive learning tasks within a single sample, making the network training independent of point cloud resolution and scale, enabling the detection of anomalies in high-resolution point clouds. PointCore \cite{zhao2024pointcore} integrates local coordinate features from the memory bank with global PointMAE features, avoiding the high computational complexity and feature mismatch issues caused by multiple memory banks. This reduces computational overhead during the inference stage. Additionally, it proposes a ranking-based normalization method to mitigate the impact of extreme values, better balancing the coordinate anomaly scores and PointMAE anomaly scores with distribution differences. R3D-AD \cite{zhou2024r3d} designs an anomaly simulation strategy for 3D point clouds, which can simulate bulge, sink, and damage on point clouds to generate pseudo samples. R3D-AD uses a diffusion model to reconstruct the 3D point cloud pseudo samples and uses reconstruction errors to detect and localize anomalies within the point clouds.

\section{Multimodal UIAD Methods}

Multimodal UIAD methods use multiple modal data as inputs. Currently, mainstream and publicly available unsupervised multimodal anomaly detection datasets all contain two modals: RGB images and 3D point clouds. Therefore, the criteria for anomaly detection include the structure information, position information, and color information of the input samples. Compared with single-modal (RGB images or 3D point clouds) anomaly detection tasks, multimodal anomaly detection requires the processing different modal information (such as different preprocessing and feature extraction methods), and requires effective information fusion (feature fusion or result aggregation) to enhance the interaction between different modal information. The paradigm of multimodal UIAD follows the same as that of RGB UIAD, which is divided into feature embedding-based methods and reconstruction-based methods.

\subsection{Feature Embedding-based Methods}

\subsubsection{Teacher-Student Architecture Methods}

AST \cite{rudolph2023asymmetric} takes RGB images and depth maps as inputs, uses an asymmetric student-teacher network structure. The teacher network uses conditional normalization flow, and the student network is a traditional convolutional neural network. AST directly connects multimodal features as fusion features as model inputs, and uses the bijectivity of the normalizing flow to further increase the difference between the output of the teacher network for normal samples and abnormal samples, thereby further increasing the difference between the output of the teacher network and the student network for abnormal samples. MMRD \cite{gu2024rethinking} takes RGB images and depth maps as inputs, proposes a multimodal reverse distillation method, which uses a teacher network with a siamese architecture as the encoder to extract features from different modals, while achieving parameter-free fusion of different modal features and conducting multimodal information interaction within a learnable student network decoder.

\subsubsection{Memory Bank Methods}

M3DM \cite{wang2023multimodal} introduces memory banks into multimodal anomaly detection tasks for the first time, takes RGB images and 3D point clouds as inputs, and uses multimodal memory banks and OCSVM to make decisions to detect and locate anomaly areas. At the same time, M3DM proposes point feature alignment to align 3D features and 2D features, and employs contrastive learning for feature fusion to encourage the same positions of different modal features to have more corresponding information, thereby reducing interference between features caused by directly concatenation of multimodal features. CPMF \cite{cao2024complementary} takes RGB images and 3D point clouds as inputs, aggregates multimodal features and stores the aggregated fusion features in a single memory bank. Shape-Guided \cite{chu2023shape} takes RGB images and 3D point clouds as inputs, also uses multiple memory banks, SDF memory bank and RGB memory bank, and connects the two memory banks through corresponding relationships. LSFA \cite{tu2024self} takes RGB images and 3D point clouds as inputs, builds global-level and local-level dynamically updated memory banks for multimodal features to minimize the distance between normal features in multi-granularity views, thereby better distinguishing normal features from abnormal features. ITNM \cite{wang2024incremental} takes RGB images and 3D point clouds as inputs, uses an incremental training method, continuously updating the memory bank with new samples to avoid retraining and catastrophic forgetting issues, which can reduce storage memory and training time. During inference, ITNM introduces pixel position information and uses a neighborhood matching strategy to compare query features with adjacent template features in the memory bank, preventing the problem of underestimation of anomaly scores caused by incorrect position matching. CMDIAD \cite{sui2024cross} takes RGB images and 3D point clouds as inputs, refers to M3DM to build multimodal memory banks and constructs specific cross-modal distillation networks to establish mappings from one modal to another. It uses a Multimodal Training, Few-modal Inference (MTFI) pipeline, where multimodal data is employed during the training phase, and in the inference phase, when there is a modal missing, the information from another modal is used to compensate and infer. M3DM-NR \cite{wang2024m3dm} adds a two-stage denoising network based on M3DM to denoise the input training data, and proposes an aligned multi-scale point cloud feature extraction module to replace the existing farthest point sampling (FPS) to extract local point cloud features.

\subsection{Reconstruction-based Methods}

EasyNet \cite{chen2023easynet} takes RGB images and depth maps as inputs, without employing pre-trained feature extractors and memory banks, which facilitates real-time deployment. EasyNet utilizes a dual-brunch multimodal reconstruction network to separately reconstruct the inputs of the two modals. Additionally, EasyNet proposes an attention-based information entropy fusion module, which compares the information entropy of channels integrating RGB and depth features with the entropy of channels only integrating pure RGB features. This comparison determines whether to fuse depth modal features. Similar to EasyNet, DBRN \cite{bi2023dual} takes RGB images and depth maps as inputs, utilizes a dual-branch multimodal reconstruction network. DBRN employs Perlin noise to generate anomaly masks on depth maps and uses grayscale images from natural datasets to simulate anomaly textures. Additionally, DBRN introduces an importance scoring module to evaluate the importance of different modal information, and then performs weighted addition to obtain fused features. 3DSR \cite{zavrtanik2024cheating} takes RGB images and depth maps as inputs, adopts a novel anomaly simulation method on depth maps, using Perlin noise to simulate locally smooth textures, and employing a randomized affine transform to simulate subtle local changes and varied average object distances. CFM \cite{costanzino2023multimodal} takes RGB images and 3D point clouds as inputs, maps 2D features to 3D features and 3D features to 2D features through lightweight cross-modal mapping, and detects and locates anomalies by calculating the reconstruction error with the multimodal features before reconstruction. This cross-modal mapping network essentially belongs to a reconstruction network. 3DRÆM \cite{zavrtanik2024keep} extends DRÆM \cite{zavrtanik2021draem} to 3D settings, taking RGB images and depth maps as inputs. 3DRÆM proposes a novel depth map anomaly simulation strategy that follows to the natural characteristics of industrial depth data and can generate various types of anomalies.

\subsection{Multimodal Feature Fusion}

Multimodal feature fusion is a key step in multimodal UIAD tasks. According to BTF \cite{horwitz2023back}, different modal data can complement each other. Fusion of different modals can cover more types of anomalies, improve detection accuracy, and enhance the detection robustness of the model in complex environments. We categorize fusion strategies according to the stage of multimodal feature fusion: early fusion, middle fusion, late fusion, and hybrid fusion.

\begin{figure}[ht]
    \centering
    \includegraphics[width=1\columnwidth]{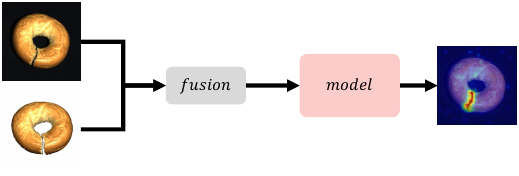}
    \caption{Early fusion strategy process in multimodal UIAD method.}
    \label{early-fusion}
\end{figure}

\subsubsection{Fusion Strategies}
\label{fusion_strategies}

\textbf{Early Fusion.} The process of early fusion is shown in Fig. \ref{early-fusion}. Early fusion refers to the fusion of different modal features before entering the main detection network (for example, after the feature extraction stage), with subsequent networks receiving only a single fused feature as input. Early fusion eliminates the need to have separate branches for different modals in the detection network, reduces the network parameters and training costs. However, the disadvantage is the difficulty in handling the overwhelming influence of a single modal, which can lead to information imbalance. AST \cite{rudolph2023asymmetric} directly concatenates RGB features and depth maps along the channel dimension as fused features before feeding them into the teacher-student network, and serves the fused features as the inputs of the subsequent network. ITNM \cite{wang2024incremental} uses pre-trained feature extractors to extract RGB features and point cloud features, and concatenates the weighted point cloud features with RGB features.

\begin{figure}[ht]
    \centering
    \includegraphics[width=1\columnwidth]{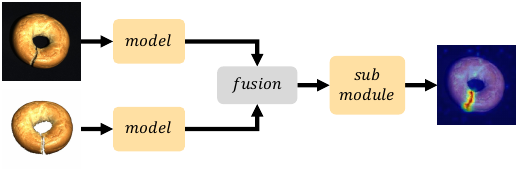}
    \caption{Middle fusion strategy process in multimodal UIAD method.}
    \label{middle-fusion}
\end{figure}

\textbf{Middle Fusion.} The process of middle fusion is shown in Fig. \ref{middle-fusion}. Middle fusion refers to the fusion of different modal features in the subject detection network and uses the fused features as the inputs of the subsequent network. Compared to early fusion, middle fusion retains separate processing networks for different modal features, and uses additional processing networks to fuse multimodal information and make decisions to obtain anomaly detection results. EasyNet \cite{chen2023easynet} does not use pre-trained feature extractors, but directly inputs RGB images and depth maps into reconstruction networks separately. EasyNet uses the features of the reconstruction process of different modals as the input of the subsequent fusion module, and obtains the anomaly detection result after an attention-based information entropy fusion module and a discriminator network. 3DSR \cite{zavrtanik2024cheating} employs channel-separated multimodal reconstruction network without multimodal feature interaction to reconstruct RGB images and depth maps respectively, and concatenates the RGB images and depth maps before and after reconstruction, feeding them into a subsequent anomaly detection module to obtain detection results. DBRN \cite{bi2023dual} is similar to EasyNet. DBRN uses a dual-branch reconstruction network to reconstruct the input RGB images and depth maps respectively, and then concatenates the RGB images and depth maps before and after reconstruction, feeding them into a subsequent discriminator network.

\begin{figure}[ht]
    \centering
    \includegraphics[width=1\columnwidth]{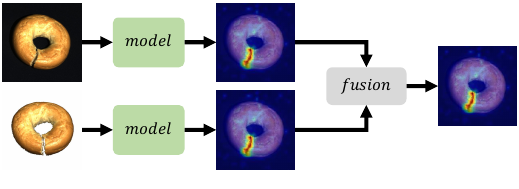}
    \caption{Late fusion strategy process in multimodal UIAD method.}
    \label{late-fusion}
\end{figure}

\textbf{Late Fusion.} The process of late fusion is shown in Fig. \ref{late-fusion}. Late fusion refers to the fusion of detection results of different modal networks after the main detection network. Late fusion discards the interaction between multimodal features, retains the independence of information from different modals and avoids information interference between modals. However, this strategy increases the model parameters and complexity, as it requires designing separate processing networks for different modals. 3DRÆM \cite{zavrtanik2024keep} extends DRÆM \cite{zavrtanik2021draem} to multimodal inputs without changing the network architecture. 3DRÆM uses the same reconstruction and discrimination networks for both RGB images and depth maps, and aggregates the final anomaly detection results to obtain the final detection results.

\textbf{Hybrid Fusion.} Hybrid fusion is a combination of the above strategies, which applies different fusion strategy at different processing stages to fully interact features across different modals, exploiting the complementary advantages of multimodal information. Compared to the above three types of fusion strategies, hybrid fusion has greater flexibility and is applicable to different types of models. Hybrid fusion is also the most commonly used multimodal feature fusion strategy. M3DM \cite{wang2023multimodal} does not directly concatenate different modal features like previous methods. Instead, M3DM aligns 3D point cloud features to the 2D plane in the feature extraction stage, and uses contrastive learning to fuse the two modal features, encouraging the same position of different modal features to have more corresponding information. M3DM also utilizes a 3D feature memory bank, a RGB feature memory bank, and a fused feature memory bank to make decisions. CPMF \cite{cao2024complementary} presents point clouds as multi-view 2D images during the feature extraction stage and extracts point cloud features and 2D features. Then CPMF concatenated the two modal features as fusion features, which are used to update the memory bank and for inference. Shape-Guided \cite{chu2023shape} uses pre-trained feature extractors to separately extract RGB features and point cloud features, utilizing point cloud patch features to guide the retrieval of RGB patch features. Shape-Guided constructs a dual-memory bank to reconstruct RGB features and point cloud features. Finally, the anomaly detection results are obtained through the combined reconstruction loss of the two modals. CFM \cite{costanzino2023multimodal} projects point cloud features onto a 2D plane to obtain 3D features. CFM extracts the relationships between features from different modal information through cross-modal mapping, and aggregates the reconstruction errors of the two modal features as anomaly detection results. LSFA \cite{tu2024self}, based on M3DM, uses contrastive learning to fuse multimodal features from local and global views. LSFA uses a RGB memory bank and a point cloud memory bank to infer the anomaly detection results respectively. The final detection results is obtained by aggregating the multimodal detection results. MMRD \cite{gu2024rethinking} uses parameter-free modal modulation in a frozen teacher network encoder to realize multimodal feature interaction and obtain fused features. MMRD then performs intra-modal interaction and inter-modal interaction in a student network decoder to obtain new fused features, and calculates pixel-wise similarity with the fused features obtained from the teacher network encoder to obtain anomaly detection results. CMDIAD \cite{sui2024cross} first trains a cross-modal distillation network to establish mappings from the feature space of one modal to the feature space of another modal, then uses memory banks to infer anomaly detection results separately for RGB features and point cloud features, and uses OCSVM to implement decision layer fusion.

\subsubsection{Fusion Methods}

Existing multimodal feature fusion methods can be divided into three categories: concatenation-based methods, projection-based methods, and learning-based methods. (1) Concatenation-based methods are the most straightforward fusion approach, where different modal features are directly concatenated along the channel dimension to obtain fused features, which are then used as input for subsequent models. (2) Projection-based methods map different modal information into a shared feature space, achieving information fusion between modals. These methods can also be combined with other fusion operations, such as concatenation after projection into the shared feature space. (3) Learning-based methods use deep learning or contrastive learning to learn how to effectively fuse features from different modals. These methods can better capture the complex relationships between modals but also face additional challenges, such as increased training costs and the risk of underfitting or overfitting.

Concatenation-based methods are often used to fuse RGB feature maps and depth maps. When using 3D point cloud features instead of depth maps, downsampling is typically required to match the size of the RGB feature map. AST \cite{rudolph2023asymmetric} downsamples both RGB image features and depth maps to the same size and concatenates them along the channel dimension to obtain the fused features. DBRN \cite{bi2023dual}, 3DRÆM \cite{zavrtanik2024keep}, and 3DSR \cite{zavrtanik2024cheating} are similar to AST, directly concatenating RGB image features and depth maps along the channel dimension and feeding them into subsequent anomaly detection (discriminative) modules. ITNM \cite{wang2024incremental} considers that the feature vectors of the two modals differ in both dimension and scale, so it applies weighting to the downsampled point cloud features to balance them before concatenating with RGB features. CPMF \cite{cao2024complementary} normalizes the RGB features and downsampled point cloud features before concatenating them along the channel dimension. Shape-Guided \cite{chu2023shape} applies an affine transformation to the RGB anomaly scores to align them with the 3D scores, then directly fuses the two modals' anomaly scores by taking the maximum score from each pixel. EasyNet \cite{chen2023easynet} calculates the information entropy of channels with only RGB features and the information entropy of channels with both RGB and depth features before concatenating the multimodal features. If the fusion of depth features enhances the information entropy, the features of both modals are then concatenated.

Projection-based methods are commonly used to fuse RGB features and 3D point cloud features. M3DM \cite{wang2023multimodal} projects the extracted 3D point cloud features onto a 2D plane using point coordinates and camera parameters, and unifies the sizes of 2D patch features and 3D patch features through average pooling. Similar to M3DM, CFM \cite{costanzino2023multimodal} and LSFA \cite{tu2024self} interpolate and project 3D point cloud features onto a 2D plane, thus enabling further fusion of features from the two modals.

Learning-based methods, compared to the previous two methods, have a broader application domain. M3DM proposes using contrastive learning to fuse RGB features and projected 3D point cloud features. It encourages multimodal patch features at the same position to have the most mutual information through patch-wise contrastive loss, meaning that the diagonal elements of the contrastive matrix have the highest values. LSFA, inspired by M3DM, uses contrastive learning to fuse multimodal features while adding an interaction of global structural information. It clusters local features to obtain instance-wise features, and then applies contrastive loss to the instance-wise features. CMDIAD \cite{sui2024cross} does not focus on information fusion between the two modals but instead establishes a mapping from the feature space of one modal to the feature space of the other modal. This results in a cross-modal distillation network that enables interaction between different modals.

\subsubsection{Fusion Strategy Optimization}

The four multimodal feature fusion strategies introduced in Section \ref{fusion_strategies} are the mainstream approaches currently used in multimodal industrial anomaly detection tasks. However, considering the high complexity of industrial scenarios and the high requirements for detection accuracy, there is still room for optimization in the existing strategies.

In terms of fusion efficiency, early fusion can reduce training costs by fusing features before detection, but it struggles to handle the overwhelming influence of a single modal, which may lead to information imbalance. Middle fusion fuses features at an intermediate stage, allowing better retention of the characteristics of each modal, but determining the optimal fusion timing is challenging and may affect feature interaction, then impacting the detection results. Late fusion combines results after detection, avoiding information interference, but it increases model parameters and processing complexity. Hybrid fusion is highly flexible, but in practical applications, the challenge remains to precisely select the fusion stage and method for different industrial scenarios, which is still an unresolved issue.

Moreover, the current strategies are difficult to adapt to certain anomaly situations in complex and dynamic industrial environments, such as missing modals. In multimodal industrial anomaly detection tasks, missing modal data may occur due to factors such as equipment failures and environmental interference. Existing multimodal detection methods require joint input from all modals, and the absence of one modal can prevent the detection process from proceeding.

3D-ADNAS \cite{long2024revisiting} optimizes multimodal anomaly detection methods from the perspective of mult-modal fusion architecture design. It introduces Neural Architecture Search (NAS), designing a two-level search space that simultaneously optimizes both intra-module (feature selection and fusion within modules) and inter-module (searching for the optimal module combination strategy) aspects. By combining gradient optimization strategies, it automatically searches for an efficient multimodal fusion architecture. 3D-ADNAS expands the perspective of multimodal fusion beyond the four previously mentioned strategies, combining existing strategies in a more flexible way to obtain more effective solutions.

RADAR \cite{miao2024radar} focuses on the problem of modal-missing in multimodal industrial anomaly detection. It introduces modality-incomplete instruction, using a pre-trained multimodal Transformer to dynamically adapt to missing modals. RADAR designs a real-pseudo label hybrid module to highlight the uniqueness of different modal combinations, enhancing anomaly localization ability, and combines a multi-feature repository (RGB, point clouds and fused features) to compute anomaly scores. RADAR effectively mitigates the impact of modal-missing cases on multimodal industrial anomaly detection tasks and presents this issue as a new task.

\begin{table*}
    \centering
    \renewcommand\arraystretch{1.0}
    \resizebox{\linewidth}{!}{
    \begin{tabular}{clclp{11cm}}
        \toprule
        & \textbf{Methods} & \textbf{Year} & \textbf{Paradigm} & \textbf{Highlight} \\
        \midrule
        \multirow{17}*{\rotatebox{90}{Few-shot}}
        & MetaFormer \cite{wu2021learning} & 2021 & Meta learning-based & Introduced meta-learning into few-shot anomaly detection for the first time. \\
        & RegAD \cite{huang2022registration} & 2022 & Meta learning-based & Designed an anomaly-free feature registration network to learn category-agnostic feature registration. \\
        & TDG \cite{sheynin2021hierarchical} & 2021 & Vanilla few-shot methods & Generated hierarchical transformations in few-shot settings. \\
        & FastRecon \cite{fang2023fastrecon} & 2023 & Vanilla few-shot methods & Proposed a regression with distribution regularization to achieve fast feature reconstruction. \\
        & GraphCore \cite{xie2023pushing} & 2023 & Vanilla few-shot methods & Introduced a vision isometric invariant feature for building a more compact memory bank. \\
        & COFT-AD \cite{liao2024coft} & 2024 & Vanilla few-shot methods & Designed a few-shot anomaly detection method based on contrastive fine-tuning. \\
        & DFD \cite{bai2024dual} & 2024 & Vanilla few-shot methods & Used frequency domain information to detect and locate image-level and feature-level anomalies in the feature space. \\
        & AnomalyGPT \cite{gu2024anomalygpt} & 2023 & Large model-based & Introduces LVLM into unsupervised industrial anomaly detection for the first time. \\
        & TGI-AD \cite{lee2024text} & 2024 & Large model-based & Applied large models to image generation tasks to reconstruct input samples. \\
        & PromptAD \cite{li2024promptad} & 2024 & Large model-based & Connected normal prompts with anomaly suffixes to construct a large number of negative samples. \\
        & InCTRL \cite{zhu2024toward} & 2024 & Large model-based & Learned between query images and normal prompts without relying on handcrafted text prompts about specific defects. \\
        \midrule
        \multirow{17}*{\rotatebox{90}{Zero-shot}}
        & WinCLIP \cite{jeong2023winclip} & 2023 & CLIP-based & Introduced the CLIP in zero-shot settings for the first time. \\
        & RWDA \cite{tamura2023random} & 2023 & CLIP-based & Used CLIP to generate training data instead of direct inference. \\
        & APRIL-GAN \cite{chen2023april} & 2023 & CLIP-based & Added additional linear layers to CLIP. \\
        & CLIP-AD \cite{chen2023clip} & 2023 & CLIP-based & Used features at different levels and applies the architecture and feature surgery strategy from CLIP Surgery. \\
        & AnomalyCLIP \cite{zhou2023anomalyclip} & 2024 & CLIP-based & Designed a generic, learnable, object-agnostic text prompt for normal and abnormal information. \\
        & FiLo \cite{gu2024filo} & 2024 & CLIP-based & Used domain-specific knowledge to introduce detailed anomaly descriptions to replace generic text prompt. \\
        & VCP-CLIP \cite{qu2024vcp} & 2024 & CLIP-based & Enhanced the anomaly semantic perception ability of CLIP through visual context prompts. \\
        & AdaCLIP \cite{cao2024adaclip} & 2024 & CLIP-based & Enhanced dynamic adaptation capabilities of model through hybrid learnable prompts. \\
        & ALFA \cite{zhu2024llms} & 2024 & CLIP-based & Introduced a run-time prompt adaptation strategy to address the issue of cross-semantic ambiguity. \\
        & SAA \cite{cao2023segment} & 2023 & SAM, DINO-based & Used GroundingDINO and SAM to detect and locate anomalies. \\
        & MuSc \cite{li2024musc} & 2024 & Vanilla zero-shot methods & Proposed a zero-shot detection model that does not require any training or textual prompts using mutual scoring. \\
        \midrule
        \multirow{14}*{\rotatebox{90}{Multi-Class}}
        & UniAD \cite{you2022unified} & 2022 & Parallel training-based & First proposed using a unified model to detect anomalies in multiple object categories. \\
        & OmniAL \cite{zhao2023omnial} & 2023 & Parallel training-based & Improved anomaly synthesis, reconstruction and localization. \\
        & RAN \cite{lu2023removing} & 2023 & Parallel training-based & Proposed a gradient denoising model to reconstruct images from multi-level noises and locate anomalies. \\
        & HVQ-Trans \cite{lu2023hierarchical} & 2023 & Parallel training-based & Proposed a VQ-based transformer and a hierarchical VQ-based approach with switching mechanism. \\
        & DiAD \cite{he2024diffusion} & 2023 & Parallel training-based & Addressed the issues of classification errors and semantic errors in existing diffusion models in multi-class settings. \\
        & LTAD \cite{ho2024long} & 2024 & Parallel training-based & Proposed long-tail anomaly detection in multi-class settings. \\
        & PNPT \cite{yao2024prior} & 2024 & Parallel training-based & Proposed semantic alignment between normal prompting and sample self-attributes. \\
        & HGAD \cite{yao2024hierarchical} & 2024 & Parallel training-based & Designed a hierarchical Gaussian mixture modeling method. \\
        & DNE \cite{li2022towards} & 2022 & Continual training-based & Introduced continual learning in the anomaly detection tasks for the first time. \\
        & UCAD \cite{liu2024unsupervised} & 2024 & Continual training-based & Addressed the inability of DNE to localize anomaly areas. \\
        & IUF \cite{tang2023incremental} & 2024 & Continual training-based & Used object category features to segregate the semantic spaces of different objects. \\
        \bottomrule
    \end{tabular}}
    \caption{A summary of learning methods regarding year, paradigm and highlight.}
    \label{tab:other_methods}
\end{table*}

\begin{table*}
    \centering
    \renewcommand\arraystretch{1.0}
    \resizebox{\linewidth}{!}{
    \begin{tabular}{clclp{11cm}}
        \toprule
        & \textbf{Methods} & \textbf{Year} & \textbf{Level} & \textbf{Highlight} \\
        \midrule
        \multirow{21}*{\rotatebox{90}{RGB}}
        & CutPaste \cite{li2021cutpaste} & 2021 & Image & Proposed a simple and efficient cut-and-paste strategy. \\
        & DRÆM \cite{zavrtanik2021draem} & 2021 & Image & Proposed a more realistic anomaly simulation strategy based on Perlin noise. \\
        & Defect-GAN \cite{zhang2021defect} & 2021 & Image & Used GAN to generate simulated anomalies. \\
        & NSA \cite{schluter2022natural} & 2022 & Image & Mitigated artificial boundary discontinuities of the simulated natural anomaly. \\
        & DFMGAN \cite{duan2023few} & 2023 & Image & Generated highly realistic anomalies by fine-tuning in the feature space. \\
        & AnomalyDiffusion \cite{hu2024anomalydiffusion} & 2024 & Image & Used diffusion models to generate anomalies with controllable appearance and location. \\
        & RealNet \cite{zhang2024realnet} & 2024 & Image & controlled anomaly intensity generated by diffusion models through perturbation parameters. \\
        & PatchAnomaly \cite{fan2024patch} & 2024 & Image & Proposed integrating self-supervised learning to synthesize patch-level anomalies. \\
        & CAF \cite{lin2024comprehensive} & 2024 & Image & Proposed a near-distribution anomaly augmentation method generated near-distribution anomaly. \\
        & CAGen \cite{jiang2024cagen} & 2024 & Image & Generated highly controllable anomalies by overlaying local perturbations on normal samples. \\
        & DMDD \cite{liu2024dual} & 2024 & Image & Extracted the foreground of RGB images and simulated anomalies only on the foreground. \\
        & MAAE \cite{liu2024mixed} & 2024 & Image & Generated adaptive noise for objects of different categories. \\
        & DSR \cite{zavrtanik2022dsr} & 2022 & Feature & Proposed simulating anomalies in discrete feature space. \\
        & SimpleNet \cite{liu2023simplenet} & 2023 & Feature & Proposed adding Gaussian noise to normal sample features. \\
        & GeneralAD \cite{strater2024generalad} & 2024 & Feature & Proposed adding noise to random locations or copy-pasting features to strongly attended regions of normal sample features. \\
        & GLASS \cite{chen2024unified} & 2024 & Image \& Feature & Combined global and local anomaly synthesis strategies. \\
        \midrule
        \multirow{5}*{\rotatebox{90}{Depth}}
        & EasyNet \cite{chen2023easynet} & 2023 & Depth & Simulated depth anomalies in the foreground area by depth thresholding. \\
        & DBRN \cite{bi2023dual} & 2024 & Depth & Simulated out-of-range depth values as protrusions and indentations by normalizing them. \\
        & 3DSR \cite{zavrtanik2024cheating} & 2024 & Depth & Simulated the anomaly using an affine transformation to simulate local depth changes. \\
        & 3DRÆM \cite{zavrtanik2024keep} & 2024 & Depth & Smoothed the depth values of simulated anomalies. \\
        \midrule
        \multirow{4}*{\rotatebox{90}{\makecell{Point-\\Cloud}}}
        & Group3AD \cite{zhu2024towards} & 2024 & Point & Simulated point cloud anomalies by adding normally distributed random noise to local area points. \\
        & R3D-AD \cite{zhou2024r3d} & 2024 & Point & Generated geometric anomalies such as bulge or sink by applying rotation, viewpoint selection, and local point transformations. \\
        \bottomrule
    \end{tabular}}
    \caption{A summary of anomaly simulation methods regarding year, level and highlight.}
    \label{tab:simulation}
\end{table*}

\section{Learning Methods}

Considering the deployment issues of anomaly detection algorithms in actual industrial production lines, some related learning methods for optimization have been proposed. In Tab. \ref{tab:other_methods}, we summarize three key optimization directions that are currently being focused on: few-shot methods, zero-shot methods, and multi-class methods. And in Tab. \ref{tab:simulation}, we summarize methods of anomaly simulation.

\subsection{Few-Shot Learning}

According to TDG \cite{sheynin2021hierarchical} and Liu et al. \cite{liu2024deep}, few-shot anomaly detection (FSAD) methods can be classified into meta learning-based methods, vanilla few-shot learning methods, and large model-based methods.

\begin{figure}[ht]
    \centering
    \includegraphics[width=1\columnwidth]{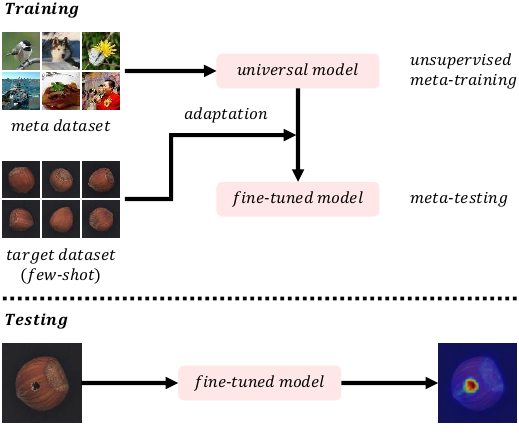}
    \caption{Process of meta learning-based methods.}
    \label{meta-learning}
\end{figure}

\subsubsection{Meta Learning-based Methods}

Meta learning-based methods require a large number of additional images as a meta-training dataset, enabling the model to quickly adapt to new tasks with a small number of samples. The process of meta learning-based methods is shown in Fig. \ref{meta-learning}. MetaFormer \cite{wu2021learning} designs a meta-learning strategy that allows general anomaly detection models to be fine-tuned with only a few normal samples when processing new categories. In the meta-training phase, MetaFormer uses additional datasets to train the model to obtain a general model, and in the meta-testing phase, it effectively updates the parameters of the general model using a few normal samples from the UIAD dataset. RegAD \cite{huang2022registration} designs an anomaly-free feature registration network to learn category-agnostic feature registration, and identifies anomalies by comparing the registration features of test samples with their corresponding normal samples under an ImageNet \cite{deng2009imagenet} pre-trained backbone.

\subsubsection{Vanilla Few-Shot Methods}

Vanilla few-shot methods do not require additional supervision and achieve FSAD through the specificity of their own models and a limited number of normal samples. TDG \cite{sheynin2021hierarchical} proposes a hierarchical transformation discriminative generative model that captures latent anomalies in images by generating hierarchical image transformations in few-shot settings. FastRecon \cite{fang2023fastrecon} proposes a regression with distribution regularization to achieve fast feature reconstruction for FSAD. GraphCore \cite{xie2023pushing} does not rely on additional datasets or pre-trained feature extractors, it employs graph neural networks to extract vision isometric invariant features, constructing more compact memory banks. COFT-AD \cite{liao2024coft} designs a FSAD method based on contrastive fine-tuning. COFT-AD achieved FSAD by fine-tuning on few-shot target dataset through contrastive training. DFD \cite{bai2024dual} starts from the frequency domain perspective, designs a dual-path frequency discriminator to detect and locate image-level and feature-level anomalies in the feature space.

\subsubsection{Large Model-based Methods.} Large model-based methods integrate the knowledge base of the large models and combines prior knowledge guided by text prompts from text encoders to calculate anomaly scores. AnomalyGPT \cite{gu2024anomalygpt} introduces Large Vision-Language Models (LVLM) into UIAD for the first time. AnomalyGPT supports multi-round dialogue, enabling context-aware few-shot learning on new datasets with strong transfer capabilities. TGI-AD \cite{lee2024text} applies large models to image generation tasks and proposes a keyword-to-prompt generator to guide the image generation process, and generates anomaly-free images similar to the input images, which are used for anomaly detection for the input image. PromptAD \cite{li2024promptad} uses a large model as the encoder and proposes semantic concatenation (SC), which connects normal prompts with anomaly suffixes to construct a large number of negative samples. This guides prompt learning in a single-class setting. Additionally, PromptAD proposes explicit anomaly margin to explicitly control the distance between normal prompt features and anomaly prompt features. InCTRL \cite{zhu2024toward} also uses a large model as the encoder to detect anomalies by learning the residuals between query images and few-shot normal sample prompts. InCTRL does not rely on handcrafted text prompts about specific defects, allowing it to applicable to popular anomaly detection tasks, including industrial defect anomaly detection, medical image anomaly detection, and semantic anomaly detection in both one-vs-all and multi-class settings.

\subsection{Zero-Shot Learning}

Zero-shot anomaly detection (ZSAD) is an extension of few-shot anomaly detection (FSAD). On the surface, ZSAD appears to be an extreme case of FSAD where the number of samples is reduced from a few to zero, but in essence, it faces different challenges. FSAD focuses on learning from a limited number of normal samples, while ZSAD focuses on how to detect and locate anomalies using existing knowledge in zero-shot settings. Existing ZSAD methods are primarily based on CLIP (Contrastive Language-Image Pre-training) \cite{radford2021learning}, using the internal knowledge of large models to discriminate anomaly areas. WinCLIP \cite{jeong2023winclip} introduces CLIP for the first time in the ZSAD task, leveraging the robust generalization capability of CLIP to perform inference for anomaly detection. RWDA \cite{tamura2023random} uses CLIP to generate training data instead of using CLIP directly for inference. APRIL-GAN \cite{chen2023april} adds additional linear layers to the CLIP model. CLIP-AD \cite{chen2023clip} uses features at different levels and applies the architecture and feature surgery strategy from CLIP Surgery \cite{li2023clip} to solve the problem of opposite predictions and irrelevant highlights that arise when directly computing text and image feature similarities using CLIP. AnomalyCLIP \cite{zhou2023anomalyclip} designs a generic, learnable, object-agnostic text prompt for normal and abnormal information, significantly enhancing the transferability of anomaly detection across various domains. FiLo \cite{gu2024filo} propose an adaptively learned fine-grained description that leverages domain-specific knowledge to introduce detailed anomaly descriptions, replacing generic text prompt for normal and abnormal information. VCP-CLIP \cite{qu2024vcp} enhances the anomaly semantic perception ability of CLIP through visual context prompts. AdaCLIP \cite{cao2024adaclip} proposes shared static prompts for all images and dynamic prompts generated for each test image, enhancing dynamic adaptation capabilities of model through hybrid learnable prompts. ALFA \cite{zhu2024llms} leverages GPT-3.5 (gpt-3.5-turbo-instruct) to generate more informative anomaly prompts and employs a contextual scoring mechansim to adaptively adjust a set of anomaly prompts on a per-image basis, thereby reducing cross-semantic ambiguity. By encoding prompt texts and images using CLIP to obtain multi-scale features, it projects global semantic alignment into the local semantic space, achieving precise anomaly localization without training or fine-tuning.

In addition to CLIP-based methods, SAA \cite{cao2023segment} uses GroundingDINO \cite{liu2023grounding} to identify the approximate location of anomalies, and then uses SAM \cite{kirillov2023segment} for detailed segmentation. MuSc \cite{li2024musc} does not require any training or text prompts, and detects and locates anomalies by mutual scoring of unlabeled images with different aggregation degrees.

\subsection{Multi-Class Learning}

Multi-class anomaly detection (MCAD) methods, also known as unified anomaly detection methods, aim to detect anomalies from different object classes with a unified model. The existing MCAD methods mainly follow two training paradigms: parallel training and continual training.

\begin{figure}[ht]
    \centering
    \includegraphics[width=1\columnwidth]{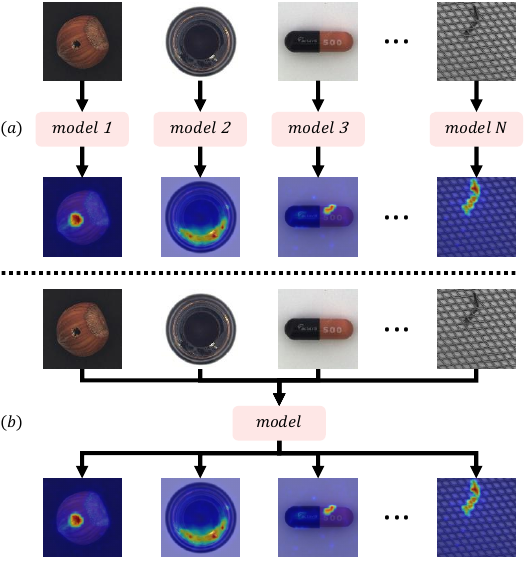}
    \caption{Comparison of the inference process between (a) separate training-based methods and (b) parallel training-based methods.}
    \label{parallel-training}
\end{figure}

\subsubsection{Parallel Training-based Methods}

The process of parallel training-based methods is shown in Fig. \ref{parallel-training}. UniAD \cite{you2022unified} proposes to use a unified model to detect anomalies from different object classes, addressing the issue where popular reconstruction networks in multi-class settings can effectively reconstruct both normal and abnormal samples (also known as “identical shortcut” issue), leading to difficulties in distinguishing anomalies. OmniAL \cite{zhao2023omnial} improves anomaly synthesis, reconstruction and localization, alleviating the model degradation problem in multi-class settings. RAN \cite{lu2023removing} uses a diffusion model for image reconstruction, training a gradient denoising model in multi-class settings to reconstruct images from multi-level noises and locate anomalies. HVQ-Trans \cite{lu2023hierarchical} extracts prototypes and propose a VQ-based (“VQ” is “Vector Quantization”) transformer, and proposes a hierarchical VQ-based approach with switching mechanism to overcome the “identical shortcut” and “prototype collapse” problem. DiAD \cite{he2024diffusion} proposes a latent-space semantic-guided network and a spatial-aware feature fusion block to address the limitations of existing diffusion models in MCAD, such as classification errors and semantic errors. LTAD \cite{ho2024long} considers that in most industrial applications, different objects have different costs, production schedules, etc., and proposes long-tail anomaly detection (i.e., the number of samples in different classes is unbalanced) in multi-class settings. PNPT \cite{yao2024prior} ensures stable reconstruction and avoids the “identical shortcut” issue through semantic alignment between normal prompting and sample self-attributes. HGAD \cite{yao2024hierarchical} designs a hierarchical Gaussian mixture modeling method to address the 'homogeneous mapping' issue in normalizing flow-based methods in multi-class settings.

\begin{figure}[ht]
    \centering
    \includegraphics[width=1\columnwidth]{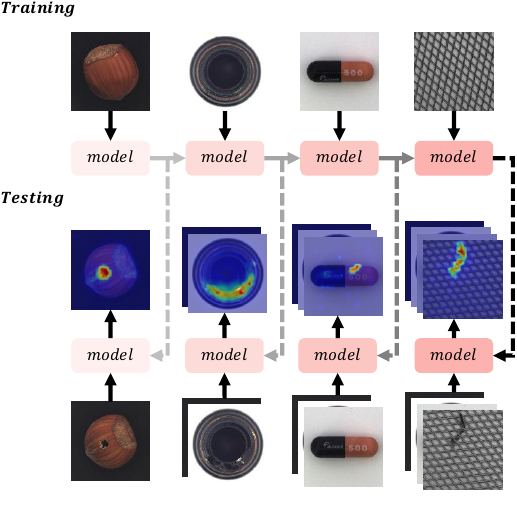}
    \caption{Process of continual training-based methods.}
    \label{continual-training}
\end{figure}

\subsubsection{Continual Training-based Methods}

Considering the sequential detection phenomenon in the actual production line, it is impossible to train all category samples at once. The process of continual training-based methods is shown in Fig. \ref{continual-training}. DNE \cite{li2022towards} introduces continual learning in the anomaly detection tasks for the first time, suitable for dynamically adjusted production lines, and effectively alleviating the catastrophic forgetting problem in the training phase of continuous anomaly detection models. However, DNE can only be used to detect anomalies but not to locate anomaly areas. UCAD \cite{liu2024unsupervised} addresses the inability of DNE to localize anomaly areas and solves the issues of catastrophic forgetting and computational burden. IUF \cite{tang2023incremental} uses object category features to segregate the semantic spaces of different objects, and prioritizes retaining the features of established objects during model weight updates, reducing the interference of new objects in the prevailing feature space, effectively solving the catastrophic forgetting problem.

\subsection{Anomaly Simulation Learning}

In industrial settings, real anomaly samples are often extremely rare, making it difficult to collect anomaly samples for model training. Simulating anomalies on normal samples to generate pseudo samples with simulated anomalies is an effective method to mitigate the scarcity of anomaly samples and address data imbalance. We classify existing anomaly simulation paradigms into sample-level (images, depth maps and point clouds, etc.) simulation and feature-level simulation. Sample-level anomaly simulation is more intuitive and can expand the training set, enhancing model robustness, but it is more challenging to simulate near-in-distribution anomalies \cite{zavrtanik2022dsr}. Features are high-dimensional data with richer semantic information, and simulating anomalies in the high-dimensional feature space is more flexible, effectively improving the model performance in detecting near-in-distribution anomalies, although it is more dependent on the feature extractor.

\subsubsection{RGB Image}

For RGB UIAD tasks, simulating anomalies in RGB images is a critical step. CutPaste \cite{li2021cutpaste} generates anomalies through simple cut-and-paste operations. Specifically, it cuts a sub-region of the image and pastes it in another location, creating discontinuous areas to obtain pseudo samples, which is simple and efficient. DRAEM \cite{zavrtanik2021draem} adopts a more realistic anomaly simulation strategy, generating anomaly areas using Perlin noise and filling textures with additional image data to create pseudo samples at the image level. Defect-GAN \cite{zhang2021defect} synthesizes high-fidelity anomaly images using a generative adversarial network (GAN). DSR \cite{zavrtanik2022dsr} simulates anomalies in the discrete feature space of normal samples to generate pseudo samples, closely matching the distribution of real anomalies and avoiding dependence on external datasets. NSA \cite{schluter2022natural} uses Poisson image editing techniques to generate anomalies by seamlessly blending enlarged local areas within the image, creating samples that simulate natural anomalies while avoiding artificial boundary discontinuities. SimpleNet \cite{liu2023simplenet} directly adds Gaussian noise to the features of normal samples to obtain anomaly features. GeneralAD \cite{strater2024generalad} distorts features by adding noise to random locations or by copy-pasting features to strongly attended regions to generate anomaly features. DFMGAN \cite{duan2023few} proposes a defect-aware feature manipulation method to generate highly realistic anomalies by fine-tuning in the feature space. AnomalyDiffusion \cite{hu2024anomalydiffusion} introduces two independent components based on diffusion models, anomaly embedding and spatial embedding, to control the appearance and location of anomalies, respectively. RealNet \cite{zhang2024realnet} generates diverse and near-normal anomaly areas using intensity-controllable diffusion models, adjusting perturbation parameters to control anomaly intensity and generate more realistic pseudo samples. PatchAnomaly \cite{fan2024patch} selects multi-scale local patches from normal images and applies self-supervised learning data augmentation techniques (such as rotation, jigsaw, and context recovering) to these patches to synthesize anomaly samples. To prevent the model from making shortcut predictions by analyzing edge information of patches, PatchAnomaly also employs sub-patch transformation and image blending. CAF \cite{lin2024comprehensive} generates simulated anomalies by classifying and selecting different augmentation methods. The simulated anomalies are categorized into transparent and opaque types, and a near-distribution anomaly augmentation (NDAA) method is proposed to generate near-distribution anomaly. CAGen \cite{jiang2024cagen} overlays local perturbations on normal samples to create highly controllable anomalies while adjusting noise intensity to control anomaly saliency. GLASS \cite{chen2024unified} combines global (adding Gaussian noise in the feature space and using gradient ascent to simulate anomalies) and local (generating stronger anomalies at the image level) anomaly synthesis strategies. DMDD \cite{liu2024dual} employs the GrabCut \cite{rother2004grabcut} algorithm to extract foreground information from RGB images and simulates anomalies only on the foreground. MAAE \cite{liu2024mixed} dynamically adjusts the intensity of noise based on the category features of objects and applies L2 regularization to the generated noise to prevent excessive noise from hindering the ability of models to accurately reconstruct the image.

\subsubsection{Depth Image}

Depth maps are similar to RGB images but have only a single channel. Existing anomaly simulation methods primarily focus on optimizing the depth values in anomaly areas. EasyNet \cite{chen2023easynet} generates a positive and negative mask map by binarizing noise images created with Perlin noise, applies the noise to the original depth images, and simulates anomalies in the foreground by generating a foreground mask. DBRN \cite{bi2023dual} creates an anomaly location mask using a Perlin noise generator and normalizes depth values within foreground areas to a specific range, with values outside this range representing protrusions and depressions as anomalies. 3DSR \cite{zavrtanik2024cheating} uses Perlin noise to generate noise maps, employing random affine transformations to simulate subtle local variations and changes in average target distance. 3DRÆM \cite{zavrtanik2024keep} uses a Perlin noise generator to create anomaly areas and smooths the simulated depth values, ensuring more consistent local depth changes.

\subsubsection{3D Point Cloud}

3D point clouds are currently the most commonly used data type for 3D UIAD tasks, but due to the complex structure of point cloud data, only a few anomaly simulation strategies exist. Group3AD \cite{zhu2024towards} selects points with the highest FPFH feature values as the central points of local areas. Around each central point, a random proportion of neighboring points is chosen to define the local area. Based on this area, random noise following a normal distribution is added to the point cloud to generate anomaly points. R3D-AD \cite{zhou2024r3d} applies random rotations to the input point cloud and then randomly selects a subset of points near a specific viewpoint on the point cloud surface to form a local area. Transformation operations are performed on these points, generating anomalies such as bulge or sink by adjusting their positions.

\section{Major Challenges}

\subsection{RGB UIAD}

\textbf{1) Difficult to promote in diverse industrial scenarios.} Currently, RGB anomaly detection methods are developing rapidly, with a large number of lightweight models, multi-class detection models, zero-shot models emerging, meeting some basic requirements for industrial deployment. However, facing the diverse business scenarios in the industrial field, most models have poor portability and are difficult to promote to multiple scenarios.

\textbf{2) Difficult to cope with complex working conditions in industrial scenarios.} RGB UIAD uses RGB images as the only criterion, but due to the influence of complex field conditions (such as uneven lighting, sampling angle deviation, background interference, etc.) and objects to be detected themselves (such as normal surface texture, different materials, etc.), the model is prone to higher false negative rate and false positive rate.

\textbf{3) Difficult to detect small-scale and subtle anomalies.} The size of anomalies on the surface of large industrial castings is significantly smaller than the size of anomalies in most publicly public RGB UIAD datasets. Small-scale and subtle anomalies (such as tiny cracks, slight scratches, or slight color differences) are not distinctive enough in features. Moreover, the image resolution is limited by the input requirements of most models, and general resolutions make it difficult to better reflect such anomalies.

\subsection{3D UIAD}

\textbf{1) Difficult to balance input formats for 3D information.} 3D anomaly detection methods mainly use 3D point clouds or depth maps (depth information extracted from point clouds, i.e., the z coordinate) as inputs. 3D point clouds provide complete spatial information, which facilitates more precise anomaly detection, but a large amount of point information requires higher data processing and computing costs. Depth information simplifies data complexity, helping the application of 2D image processing techniques to 3D information, but it results in the loss of some spatial information.

\textbf{2) Negative effects of existing 3D pre-trained feature extractors.} Most existing anomaly detection methods based on 3D point clouds rely on pre-trained feature extractors, but these pre-trained models are usually trained on large datasets and are not entirely suitable for anomaly detection tasks in specific scenarios.

\subsection{Multimodal UIAD}

\textbf{1) Difficult to solve the problem of missing modals and data noise.} Existing multimodal industrial anomaly detection methods require all modals to provide criteria jointly. In actual industrial production lines, due to sensor failures or environmental factors, data from some modals may be missing or contain noise. Additionally, considering the complexity of real-world production lines, other modals beyond RGB images and 3D point clouds may also be required, such as infrared images, grayscale images, and other monochrome images.

\textbf{2) Difficult to fully align and effectively fuse multimodal data.} Existing public multimodal datasets provide one-to-one corresponding multimodal data. In actual industrial production lines, errors in the collection angles of acquisition devices may result in multimodal data not being perfectly aligned.

\textbf{3) Difficult to deploy on actual production lines.} Multimodal UIAD is essentially the same as 2D UIAD. It is part of the industrial quality inspection process and ultimately needs to be deployed in an industrial environment. Therefore, deployment issues on actual production lines should be considered.

\section{Future Prospects}

\subsection{RGB UIAD}

Develop anomaly detection models with strong transferability and generalization to adapt to different industrial scenarios and reduce the need for customization. Improve the robustness and denoising capabilities of the algorithms to address complex working conditions and the diversity of object surfaces, and reduce false negative rate and false positive rate. Research algorithms that are capable of processing high-resolution images or capturing tiny features to improve the detection capability of small-scale and subtle anomalies.

\subsection{3D UIAD}

Research more efficient data processing methods that can handle large-scale point cloud data, or adopt strategies that combines point cloud with depth information to reduce computing costs and compensate for data loss. Develop pre-trained feature extractors for specific scenarios, or reduce dependence on existing pre-trained models to improve the accuracy and applicability of 3D anomaly detection.

\subsection{Multimodal UIAD}

Optimize algorithms to compensate for information gaps caused by modal missing or noise, enhancing denoising and anti-interference capabilities. Consider incorporating more modals into the dataset and methods, such as infrared images, grayscale images, and others, to create a unified dataset that includes multiple modals such as RGB, 3D, and infrared. Methods based on this dataset can select specific modals for anomaly detection. Research algorithms capable of aligning different modal data and effectively achieving multimodal information fusion to address the issue of incomplete alignment of multimodal data. In multimodal detection methods, consider optimization directions such as lightweight models, real-time inference, multi-class detection, few-shot detection, and zero-shot detection to improve deployment effect on actual production lines.

\section{Conclusion}

In this paper, we provide a comprehensive review of the latest progress in unsupervised industrial image anomaly detection in RGB, 3D, and multimodal settings. Based on the concept, datasets, and evaluation metrics of unsupervised industrial image anomaly detection, we review and summarize the research from the perspective of different modal settings, and classify the multimodal feature fusion strategies in multimodal setting. Additionally, considering the algorithm deployment in practical production lines, we summarize methods such as few-shot learning, zero-shot learning, multi-class learning, and anomaly simulation learning. Through a comprehensive analysis, we highlight the challenges faced by anomaly detection research in different modal settings in real-world applications, providing valuable references and directions for future research. We hope that this paper can provide the latest information reference for the research on multimodal anomaly detection tasks and provide suggestions for future development directions.



\bibliographystyle{cas-model2-names}

\bibliography{reference}



\end{document}